\documentclass[final,5p,times,twocolumn]{elsarticle}

\usepackage{amssymb}
\usepackage{amsmath}
\usepackage{hyperref}

\usepackage{float}
\usepackage{graphicx}
\usepackage{amsmath,amssymb}
\usepackage{booktabs}
\usepackage{multirow}
\usepackage{colortbl}
\usepackage{pifont}  
\usepackage{caption}
\usepackage{subcaption}  
\usepackage{float}
\usepackage{adjustbox}  
\usepackage[table,xcdraw]{xcolor}
\usepackage{ulem}
\usepackage{cleveref}
\usepackage{ragged2e}
\usepackage{afterpage} 
\usepackage{placeins}
\usepackage{lineno}
\usepackage{balance}
\crefname{figure}{Fig.}{Figs.}
\Crefname{figure}{Fig.}{Figs.}
\crefname{table}{Table}{Table}
\Crefname{table}{Table}{Table}

\usepackage{soul, color, xcolor}
\definecolor{myyellow}{RGB}{255,255,0}
\sethlcolor{myyellow}  
\newif\ifhighlight     
\highlightfalse        
\newcommand{\hlfix}[1]{\ifhighlight\hl{#1}\else#1\fi}

\newif\ifhighlightB   
\highlightBfalse     
\newcommand{\hlB}[1]{\ifhighlightB{\sethlcolor{myyellow}\hl{#1}}\else#1\fi}

\soulregister{\textbf}7
\soulregister{\textit}7
\soulregister{\cite}7 
\soulregister{\cref}7 
\soulregister{\Cref}7  
\soulregister{\underline}7 %

\journal{Information Fusion}

\begin{document}

\begin{frontmatter}

\title{Relational Representation Learning Network for Cross-Spectral Image Patch Matching}


\author[a,b]{Chuang Yu}
\author[a]{Yunpeng Liu*}
\author[a,b]{Jinmiao Zhao}
\author[c]{Dou Quan}
\author[a]{Zelin Shi}
\author[d]{Xiangyu Yue*}

\address[a]{Shenyang Institute of Automation, Chinese Academy of Sciences, Shenyang, China}
\address[b]{University of Chinese Academy of Sciences, Beijing, China}
\address[c]{Xidian University, Xi'an, China}
\address[d]{MMLab, The Chinese University of Hong Kong, Hong Kong, China}



\begin{abstract}
Recently, feature relation learning has drawn widespread attention in cross-spectral image patch matching. \hlfix{However, existing related research focuses on capturing diverse feature relations between image patches and ignores sufficient intrinsic feature representations of individual image patches. To address this limitation, we propose an innovative relational representation learning that simultaneously focuses on sufficiently mining the intrinsic features of individual image patches and the feature relations between image patches}. Based on this, we construct a \hlfix{\textbf{\underline{R}}elational \textbf{\underline{R}}epresentation \textbf{\underline{L}}earning \textbf{\underline{Net}}work (\textbf{RRL-Net})}. Specifically, we innovatively construct an autoencoder to effectively characterize the individual intrinsic features, and introduce a feature interaction learning (FIL) module to extract deep-level feature relations. \hlfix{Meanwhile, to further fully mine individual intrinsic features}, a lightweight multi-dimensional global-to-local attention (MGLA) module is constructed to enhance the global feature extraction of individual image patches and capture local dependencies within global features. By combining the MGLA module, we further explore the feature extraction network and construct an attention-based lightweight feature extraction (ALFE) network. \hlfix{Furthermore, a multi-loss post-pruning (MLPP) optimization strategy is proposed, which can greatly facilitate network optimization while avoiding increases in parameters and inference time}. Extensive experiments demonstrate that our RRL-Net achieves state-of-the-art (SOTA) performance on multiple public datasets. Our code \hlfix{is} available at \url{https://github.com/YuChuang1205/RRL-Net}
\end{abstract}

\begin{keyword}
Cross-spectral image patch matching \sep Image intrinsic features \sep Relational representation learning



\end{keyword}

\end{frontmatter}



\section{Introduction}
\label{introduction}
Image patch matching is to determine whether image patches originate from the same interest point or have the same identity. It is widely used in image registration~\cite{cao2023pcnet,wang2018deep,quan2022self,quan2022deep}, image fusion~\cite{kim2016joint,jia2023multiscale}, image retrieval~\cite{shi2022beyond,lin2023zero}, multi-view stereo reconstruction~\cite{seitz2006comparison,luo2019p} and other fields. Different from object detection and object segmentation tasks that require the precise location of object boundaries~\cite{yu2022pay,he2022cross,zhou2022generalizable,yu2022precise}, image patch matching focuses on measuring the similarity between image patches~\cite{quan2021multi,yu2023feature}. Considering that single-spectral images have limitations, cross-spectral image matching can establish the correspondence between different spectral images, providing a basis for \hlfix{making full use of complementary information}~\cite{aguilera2016learning,ofir2018deep,quan2019afd,quan2021element,yu2022multibranch}. Compared with single-spectral image patch matching, cross-spectral image patch matching not only faces illumination changes and geometric changes, but also needs to overcome pixel-level nonlinear differences between cross-spectral image patches~\cite{yu2023feature,yu2023efficient}.

Early cross-spectral image patch methods are mainly inspired by research on single-spectral image patches~\cite{lowe2004distinctive,dalal2005histograms} to construct non-deep learning methods~\cite{firmenichy2011multispectral,pinggera122012cross,aguilera2012multispectral,mouats2014multispectral,aguilera2015lghd}. However, non-deep learning-based methods have limited feature extraction capabilities. Subsequently, deep learning-based methods have gradually become a trend. Early deep learning-based cross-spectral image patch matching methods use individual image patch features for discrimination~\cite{aguilera2016learning,en2018ts,aguilera2017cross}. Considering that this task is to measure the similarity between image patches, the relation learning between image patch features is better than the learning of individual image patch features~\cite{quan2021multi,yu2023feature,yu2023efficient}. Therefore, subsequent deep learning-based methods~\cite{yu2023feature,quan2019afd,quan2021element,yu2022multibranch,yu2023efficient,quan2018cross,ko2023spectral} focus on \hlfix{extracting diverse feature relations between image patches}. However, sufficient representation of the individual image patch intrinsic features is the basis for subsequent mining of feature relations. \hlfix{Only focusing on diverse feature relations between image patches limits the improvement of matching performance.}

To solve this problem, an innovative relational representation learning is proposed for the first time, which simultaneously focuses on \hlfix{both the full mining of the intrinsic features of individual image patches and the feature relations between image patches. Intrinsic feature mining emphasizes extracting rich and discriminative internal characteristics such as semantic content and structural information from individual image patches, while feature relation learning emphasizes modeling diverse and deep-level feature relations between image patches within a shared representation space.} This idea breaks the bottleneck of subsequent feature relation extraction caused by insufficient intrinsic feature mining in existing methods and provides a new exploration direction for subsequent research on methods based on feature relation learning. Specifically, we \hlfix{design} an innovative autoencoder. The intrinsic features of individual image patches are fully characterized by using self-supervised learning. Meanwhile, a feature interaction learning (FIL) module is introduced to interactively learn the intrinsic features of image patches to \hlfix{mine richer feature relations}.

In addition, \hlfix{considering the small size of image patches and the goal of capturing overall similarity rather than pixel-level correspondence, global contextual features should be given more attention alongside local representations. To this end, we propose a multi-dimensional global-to-local attention (MGLA) module that can efficiently extract global features from multiple dimensions while capturing local dependencies within them. The parameters of the MGLA module are negligible. Building on the MGLA module, an efficient attention-based lightweight feature extraction (ALFE) network is constructed that delivers strong feature extraction with minimal computational overhead, making it well-suited for cross-spectral image patch matching.}

Furthermore, considering that reasonably strong supervision can promote matching network optimization, we propose a multi-loss post-pruning (MLPP) optimization strategy, which not only has the advantage of promoting network optimization by adding feature reconstruction branches and multiple metric networks, but also avoids the disadvantages of increasing parameters and reducing inference speed.

\begin{figure}[t]
	\centering
	\includegraphics[width=\columnwidth]{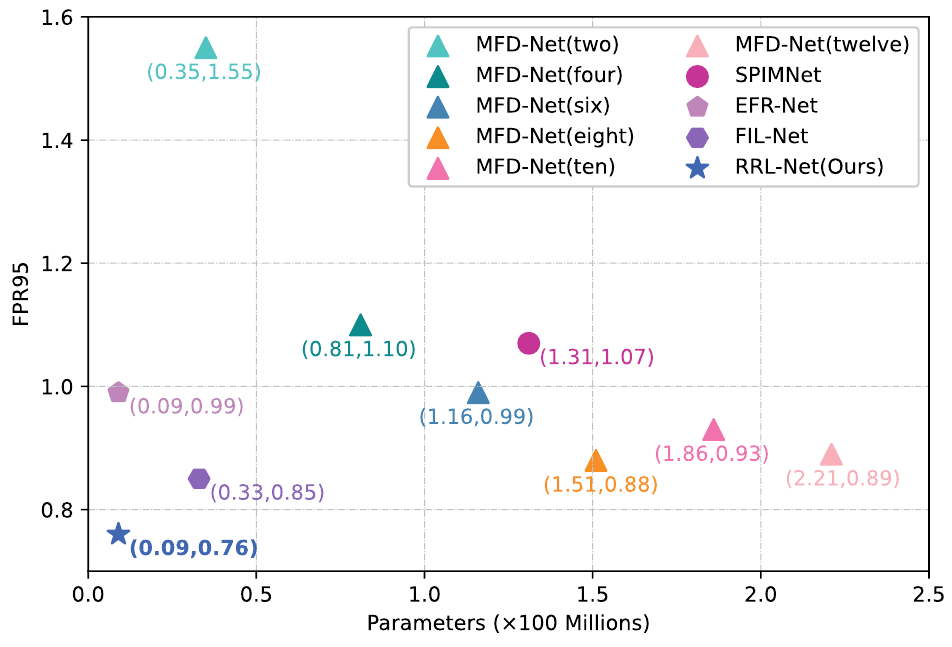}
	\vspace{-19pt}
	\caption{\hlfix{Performance comparison of multiple excellent cross-spectral image patch matching networks on the VIS-NIR patch dataset. The closer to the origin, the better the network performance. The number in parentheses for MFD-Net denotes its branch count. Our RRL-Net achieves SOTA performance. Compared with the lightweight EFR-Net~\cite{yu2023efficient}, RRL-Net has the same parameters and reduces the FPR95 by \textbf{23.2\%} (from 0.99 to 0.76). Compared with the latest and most competitive FIL-Net~\cite{yu2023feature}, RRL-Net reduces the FPR95 by \textbf{10.6\%} (from 0.85 to 0.76) and reduces parameter count by \textbf{72.7\%} (from 0.33 to 0.09).}}
	\label{fig:point_fig}
	\vspace{-7pt}
\end{figure}

In summary, we propose a Relational Representation Learning Network (RRL-Net) to fully characterize the effective feature relations between image patches. From \cref{fig:point_fig}, our RRL-Net has achieved significant performance improvements in both matching results and parameters. Our contributions can be summarized as follows:
\begin{itemize}
    \item An innovative relational representation learning is proposed, which breaks the bottleneck of subsequent feature relation extraction caused by insufficient individual intrinsic feature mining in existing methods.
    \item To further fully mine individual intrinsic features, a lightweight MGLA module is proposed and an efficient ALFE network is built based on this module.
    \item \hlfix{A MLPP optimization strategy is proposed that can promote network optimization without introducing additional parameters or increasing inference time.}
    \item \hlfix{Extensive experiments show that our RRL-Net achieves SOTA performance on multiple public cross-spectral datasets and has excellent robustness and generalization.}
\end{itemize}

\section{Related Work}
\label{related-work}
In this section, we will briefly review single-spectral image patch matching methods, cross-spectral image patch matching methods, and attention mechanisms.

\subsection{Single-spectral Image Patch Matching}
Early single-spectral image patch matching methods are mainly non-deep learning methods, including methods based on image grayscale information~\cite{barnea1972class,azriel1977coarse,vanderbrug1977two,viola1997alignment} and methods based on handcrafted feature descriptors~\cite{lowe2004distinctive,dalal2005histograms,ke2004pca,bay2008speeded,rublee2011orb,leutenegger2011brisk,alcantarilla2012kaze,alcantarilla2011fast}. However, non-deep learning-based methods do not perform satisfactorily in complex image change scenarios. \hlfix{With advances in artificial intelligence, deep learning-based methods have emerged as a prevailing research trend and can be categorized into descriptor learning and metric learning. Descriptor learning methods extract high-level features (descriptors) from individual image patches and measure similarity by computing the distance between descriptors.} Existing research on descriptor learning methods has focused mainly on improving loss functions~\cite{simo2015discriminative,balntas2016learning,balntas2016pn,kumar2016learning,zhang2017learning,yang2017deepcd,wang2019better,tian2019sosnet,tian2020hynet} and sampling strategies~\cite{tian2017l2,mishchuk2017working,wang2017train,zheng2019hardness}. For the research on loss functions, it mainly includes hinge embedding loss~\cite{simo2015discriminative} and triplet loss~\cite{balntas2016learning,balntas2016pn}. Unlike the hinge embedding loss, which uses absolute distance, the triplet loss uses relative distance. Since triplet loss has better stability, subsequent research has mainly combined triplet loss and studies of its variants~\cite{kumar2016learning,zhang2017learning,yang2017deepcd,wang2019better,tian2019sosnet,tian2020hynet}. For the research on sampling strategies, L2-Net~\cite{tian2017l2} proposes a progressive sampling strategy \hlfix{to generate numerous negative samples. However, most of these are simple negative samples that fail to provide sufficient supervisory signals during training.} Subsequently, fully mining hard negative samples has been widely studied~\cite{mishchuk2017working,wang2017train,zheng2019hardness}. \hlfix{However, overemphasizing hard negative samples and ignoring the rich information of simple samples can cause training instability or even model collapse. Different from descriptor learning methods, metric learning employs end-to-end deep networks that focus on optimizing network architecture.} MatchNet~\cite{han2015matchnet} is one of the earliest metric learning methods for image patch matching. \hlfix{Subsequently, ~\cite{zagoruyko2015learning} proposes and validates that a 2-channel network, which computes feature similarity in the fused feature space, outperforms Siamese and Pseudo-Siamese networks that directly compute similarity after extracting features in separate spaces.} Inspired by this, MRAN~\cite{quan2021multi} achieves excellent matching performance by extracting various feature relations between image patches.

\subsection{Cross-spectral Image Patch Matching}
Compared with single-spectral image patch matching, research on cross-spectral image patch matching is relatively late, and its early research is deeply inspired by single-spectral image matching methods. Non-deep learning methods for cross-spectral image patch matching mainly include reducing the impact of gradient direction changes between cross-spectral images~\cite{firmenichy2011multispectral,pinggera122012cross} and making full use of the strong correlation of edges between cross-spectral images~\cite{aguilera2012multispectral,mouats2014multispectral,aguilera2015lghd}. However, these non-deep learning-based methods are difficult to fully extract the discriminative features of cross-spectral image patches. Inspired by~\cite{zagoruyko2015learning},~\cite{aguilera2016learning} proposes Siamese networks, Pseudo-Siamese networks, and 2-channel networks suitable for cross-spectral image patch matching tasks. Compared with non-deep learning methods, all three networks have achieved significant improvement. Subsequently, SCFDM~\cite{quan2018cross} proposes to extract semantically invariant features between cross-spectral image patches in a shared semantic feature space and achieves excellent results, which further reveals that the exploration of feature relations between cross-spectral image patches is superior to the feature extraction of individual image patches. Subsequently, AFD-Net~\cite{quan2019afd,quan2021element} uses feature difference learning to extract differential features between image patches. However, feature difference learning will lose some useful discriminative features while reducing redundant interference features. Therefore, MFD-Net~\cite{yu2022multibranch} enhances the limited discriminative feature extraction capability of AFD-Net by reasonably constructing a multi-branch network structure. SPIMNet~\cite{ko2023spectral} utilizes cross-spectral image patch mutual conversion and feature difference learning to extract effective discriminative features. EFR-Net~\cite{yu2023efficient} rationally uses diverse feature relations to effectively mine the differential features and consistent features between cross-spectral image patches with low resource consumption. Although feature relation learning methods such as feature difference learning have achieved excellent results, they can only extract shallow feature relations between image patches. Therefore,~\cite{yu2023feature} proposes a feature interaction learning and constructs the FIL-Net, which can extract richer and deeper feature relations. \hlfix{However, the above methods based on feature relation learning primarily focus on extracting diverse inter-patch feature relations, while overlooking the sufficient representation of intrinsic features within individual image patches. As the foundation for effective relation modeling, such intrinsic feature should be adequately learned. Therefore, we explore sufficient representation learning of the individual intrinsic features and the feature relations between image patches.}

\definecolor{darkred}{RGB}{230,0,0}   
\definecolor{darkgreen}{RGB}{0,230,0}   
\begin{figure*}[tb]
	\centering
	\includegraphics[width=0.9\textwidth]{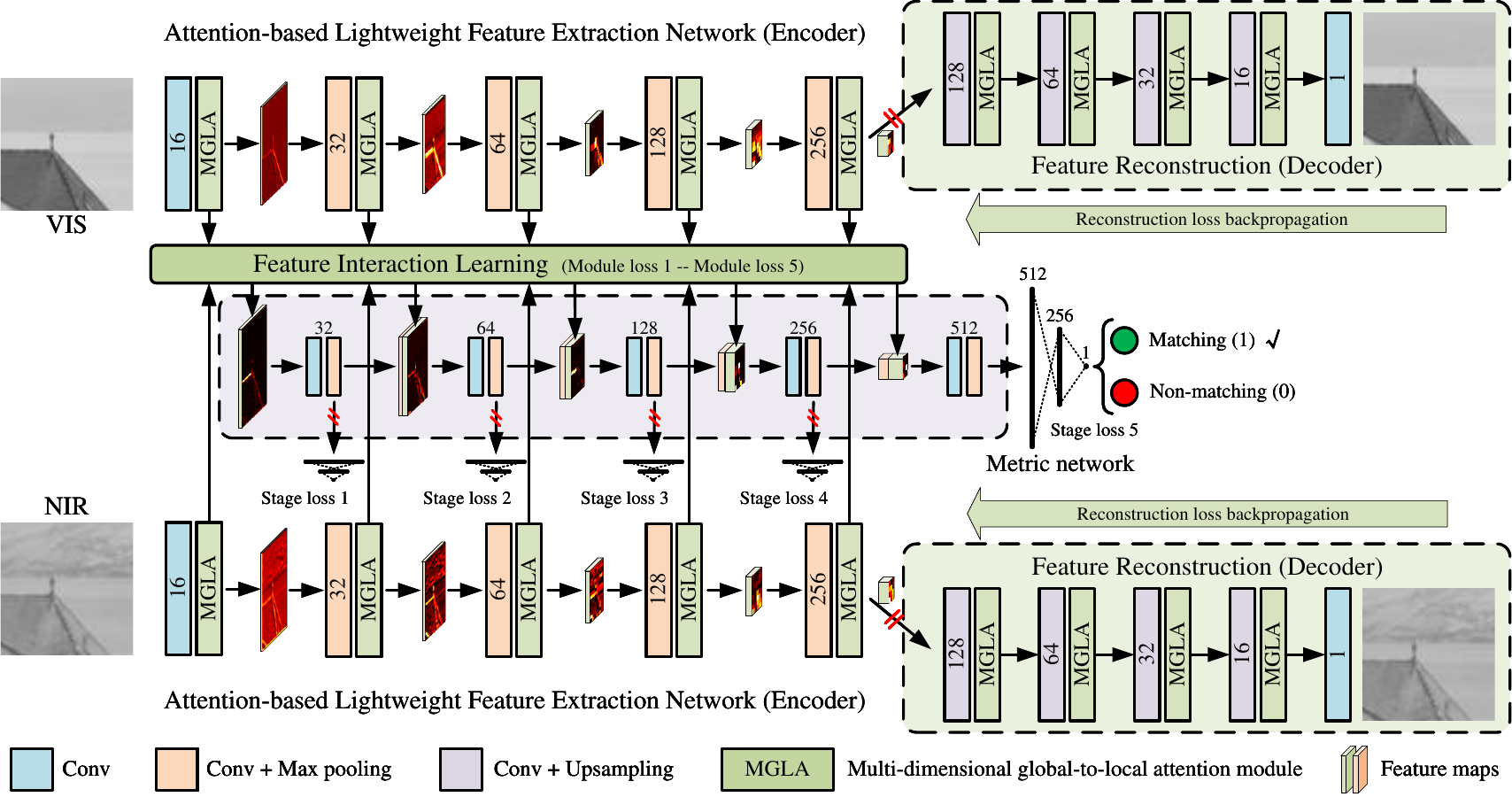}
	\vspace{-2pt}
	\caption{Network structure of RRL-Net. The \textcolor{darkred}{\textit{red}} slash denotes the post-pruning operation.}
	\label{fig:rrl-net-main}
	\vspace{-2pt}
\end{figure*}

\subsection{Attention Mechanism}
Inspired by the human visual system, the attention mechanism focuses on understanding and extracting key useful information. Generally, neural networks provide implicit attention to extract useful features from data~\cite{hassanin2024visual}. To enhance the extraction of key features, rationally adding attention mechanisms to neural networks has become a positive trend. A large amount of existing research explores various effective attention structures to improve the performance of deep learning networks on tasks such as machine translation~\cite{vaswani2017attention}, visual recognition~\cite{wang2018non} and generative models~\cite{zhang2019self}.

Vision-based attention mechanisms can be roughly divided into channel attention, spatial attention and self-attention. Channel attention considers the importance of each channel~\cite{yu2023efficient,hu2018squeeze,wang2020eca,zhang2022resnest}, such as the SE module~\cite{hu2018squeeze} and ECA module~\cite{wang2020eca}. Similar to channel attention, spatial attention aims to capture the importance of different regions in the spatial dimension~\cite{jaderberg2015spatial,woo2018cbam,zhao2019pyramid,li2020spatial,huang2019ccnet}, such as the CBAM module~\cite{woo2018cbam} and SAP module~\cite{li2020spatial}. Self-attention obtains larger receptive fields and contextual information by capturing global information~\cite{wang2018non,cao2019gcnet,fu2019dual,johnson2016perceptual,simonyan2014very}, such as Non-local block~\cite{wang2018non} and GC block~\cite{johnson2016perceptual}. Although the above attention mechanisms have achieved great performance improvements in corresponding tasks, there are few existing studies that use attention mechanisms in cross-spectral image patch matching. Therefore, combined with the characteristics of the cross-spectral image patch matching task, we attempt to build an efficient attention mechanism to fully extract the global features of image patches and the local dependencies within the global features.

\section{Method}
\label{method}
\subsection{Overall Architecture}
To efficiently extract discriminative features between cross-spectral image patches, we propose a lightweight RRL-Net to fully characterize the feature relations between them. The \cref{fig:rrl-net-main} shows the overall architecture of RRL-Net, which consists of five parts: feature extraction (encoder), feature reconstruction (decoder), feature interaction, feature aggregation and feature metric. \hlfix{For the feature extraction part, we introduce a lightweight MGLA module and further construct an efficient ALFE network specifically designed for the feature extraction in cross-spectral image patch matching.} For the feature reconstruction part, we build a feature reconstruction branch, which forms an autoencoder with the ALFE network. The feature reconstruction branch will drive the ALFE network to fully mine individual intrinsic features through self-supervised learning. It is conducive to the subsequent extraction of rich feature relations. \hlfix{Notably, the feature reconstruction branch is pruned before the model is saved to improve the inference speed.} For the feature interaction part, we introduce the FIL module, which was proposed in our previous work~\cite{yu2023feature}. It can effectively extract common features and private features between cross-spectral image patch features. In addition, we impose reasonable constraints on the common features output by the FIL module to promote the discriminability of the private features to be aggregated. For the feature aggregation part, we aggregate the private features output by the FIL module and learn again to improve feature discriminability. Its detailed structure is shown in the purple area of \cref{fig:rrl-net-main}. For the feature metric part, it contains all metric networks. Specifically, we perform a global average pooling on the feature map to convert it into a one-dimensional (1D) vector and measure the similarity through a 3-layer fully connected network, containing 512, 256 and 1 neurons, respectively. Except for the final metric network, which is used to output the similarity, other metric network branches are pruned before the network model is saved.

\hlfix{The proposed RRL-Net is substantially different from our previous work FIL-Net~\cite{yu2023feature}. Compared with FIL-Net, RRL-Net provides a more systematic and unified framework design. Their detailed differences are as follows: 1. RRL-Net proposes a new paradigm. Unlike FIL-Net which focuses on mining the diverse feature relations between image patches, RRL-Net decouples and emphasizes the modeling of individual intrinsic features and inter-patch feature relations. 2. A self-supervised autoencoder is introduced to enhance the feature representation of individual image patches. 3. RRL-Net pays greater attention to capturing the global information of image patches. A lightweight MGLA module is constructed to model the spatial information of image patches from multiple perspectives, improving the network's ability to model global semantics and local details. 4. In addition to improving matching accuracy, RRL-Net is more focused on the lightweight design. Based on the proposed MGLA module, we further construct an efficient ALFE network to extract features. At the same time, a MLPP optimization strategy is designed, which takes into account both matching accuracy and computational efficiency.}

\subsection{Relational Representation Learning}
The goal of cross-spectral image patch matching is to measure the similarity between image patches.  \hlfix{Compared to learning individual patch features in isolation, modeling the feature relations between image patches has shown superior effectiveness~\cite{quan2021multi,yu2023feature,yu2023efficient}. However, existing relation learning methods often emphasize the diversity of inter-patch relations while overlooking the sufficient representation of intrinsic features within individual patches. Since robust intrinsic representations form the foundation for accurate relation modeling, we propose a relational representation learning that focuses on fully mining two aspects: the intrinsic features of individual image patches and the feature relations between image patches.}

To fully mine the intrinsic features of individual image patches, on the one hand, we conduct an in-depth exploration of the feature extraction network and propose an ALFE network, which has efficient feature extraction capabilities and outputs five-stage feature maps. On the other hand, we reconstruct the high-level feature maps generated by the ALFE network using a feature reconstruction network (decoder). By constructing the autoencoder and using self-supervised learning, it is ensured that the ALFE network as the encoder can more fully characterize individual image patch intrinsic features.

\hlfix{To thoroughly explore the feature relations between image patches, we introduce the FIL module~\cite{yu2023feature}, which can effectively capture richer and deeper feature relations between cross-spectral image patches. Specifically, the FIL module jointly mines common and private features between image patches through enhanced feature interactions, while imposing constraints on common features to improve the discriminability of private features.} To fully \hlfix{capture} the multi-scale feature relations between image patches, we use five FIL modules on the corresponding feature maps of each stage output by the ALFE network and output five-stage multi-scale private features. At the same time, we use feature aggregation operations to sequentially aggregate multi-stage private features to extract multi-scale discriminative features.

\subsection{MGLA Module}
For an image patch with a size of only 64×64 pixels, the effective local features that can be extracted are limited. Convolution is local perception. A feature extraction network constructed by simply stacked convolution will extract a large number of invalid or repeated local features. Although the global features of image patches can be focused on by stacking multiple layers of convolution, it is far from enough. At the same time, unlike tasks such as object segmentation that require precise positioning of object boundaries, matching tasks focus on whether images match. Fully considering the global features of image patches will be more helpful in extracting richer effective features. \hlfix{Therefore, we propose a MGLA module and embed it into the feature extraction network.}

\definecolor{lightyellow}{RGB}{255,192,0}   
\definecolor{lightgreen}{RGB}{157,187,97}   
\definecolor{lightpurple}{RGB}{171,154,192}   

\begin{figure}[t]
	\centering
	\includegraphics[width=\columnwidth]{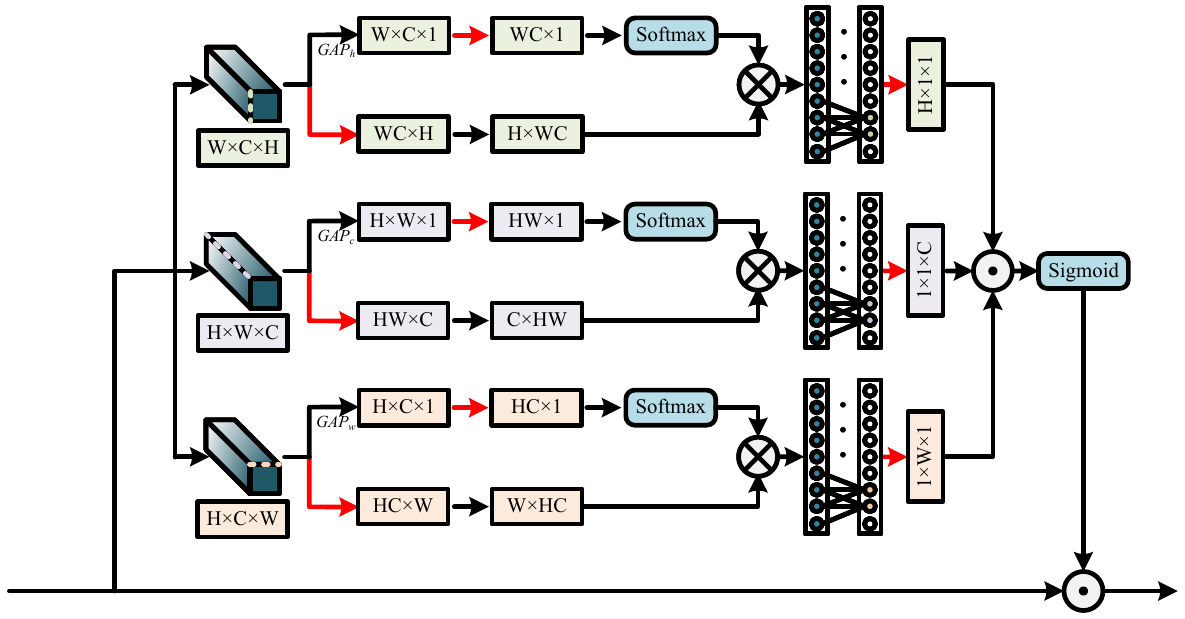}
	\vspace{-15pt}
	\caption{\hlfix{Structure of the MGLA module. $GAP$ denotes the global average pooling. $\otimes$ denotes the matrix multiplication. $\odot$ denotes the element-wise multiplication. The} \textcolor{darkred}{\textit{\hlfix{red}}} \hlfix{arrow denotes the $Reshape$ operation.} \textcolor{lightgreen}{\textit{\hlfix{Green}}}, \textcolor{lightyellow}{\textit{\hlfix{yellow}}}, \hlfix{and} \textcolor{lightpurple}{\textit{\hlfix{purple}}} \hlfix{denote operations applied to the height (\textit{H}), width (\textit{W}), and channel (\textit{C}) dimensions, respectively.}}
	\label{fig:mgla-module}
	\vspace{-5pt}
\end{figure}

From \cref{fig:mgla-module}, the proposed MGLA module performs the same operation from three different dimensions $H$, $W$ and $C$. Below, we take the channel dimension $C$ as an example to illustrate. We first extract the global context through the attention pooling operation and output a 1D global feature vector. This operation does not contain any learning parameters:
\begin{equation}
      G_c^i = DT(Re({F^i})) \otimes Softmax(Re(GA{P_c}({F^i})))
\end{equation}
where ${GAP}_c$ denotes global average pooling along the channel dimension. $Re$ denotes the $Reshape$ operation. $DT$ denotes dimension transformation.

Then, to further capture the dependencies between channels~\cite{wang2018non,cao2019gcnet} and consider the impact of SE-Net compressed channels on performance improvement~\cite{wang2020eca}, we perform local interactions within global features: 
\begin{equation}
	GL_c^i = \varphi _c^i(G_c^i)
\end{equation}
where $\varphi^i_c$ denotes the 1D convolution operation on channel dimension $C$ in the $i_{th}$ MGLA module.

The size of the convolution kernel is determined based on the length of the 1D vector, formulated as:
\begin{gather}
	k = \left\lceil {\frac{{\log _2^{len(G_c^i)} + 1}}{2}} \right\rceil   \\ 
	k = \left\{ \begin{array}{l}
				k\quad \quad \quad k\% 2 = 1\\
				k + 1\quad \;k\% 2 \ne 1
			\end{array} \right.
\end{gather}
where $len(G_c^i)$ denotes the 1D global context vector length of the $i_{th}$ stage. $\lceil\thinspace\thinspace\rceil$ denotes rounding up. $k$ denotes the 1D convolution kernel size. $\%$ denotes the remainder operation.

Each pixel generally has a stronger correlation with its surrounding pixels. To extract richer global effective features and local dependencies within global features, our operation on the channel dimension is extended to spatial dimensions:
\begin{equation}
	F_{out}^i = {F^i} \odot (Sigmoid(Re(GL_h^i) \odot Re(GL_w^i) \odot Re(GL_c^i)))
\end{equation}
where ${GL}_h^i$ and ${GL}_w^i$ denote the outputs of spatial dimensions $H$ and $W$, respectively.

In addition, combined with the MGLA module, we conduct an in-depth exploration of the feature extraction network for cross-spectral image patches. We find that the MGLA module can promote the reduction of stacked convolutions and cross-layer connections at each stage, thereby turning the constructed ALFE network into a lightweight and effective feature extraction network. Detailed experiments are provided in \cref{sec:ablation-study}.

\begin{figure*}[tb]
	\centering
	\includegraphics[width=\textwidth]{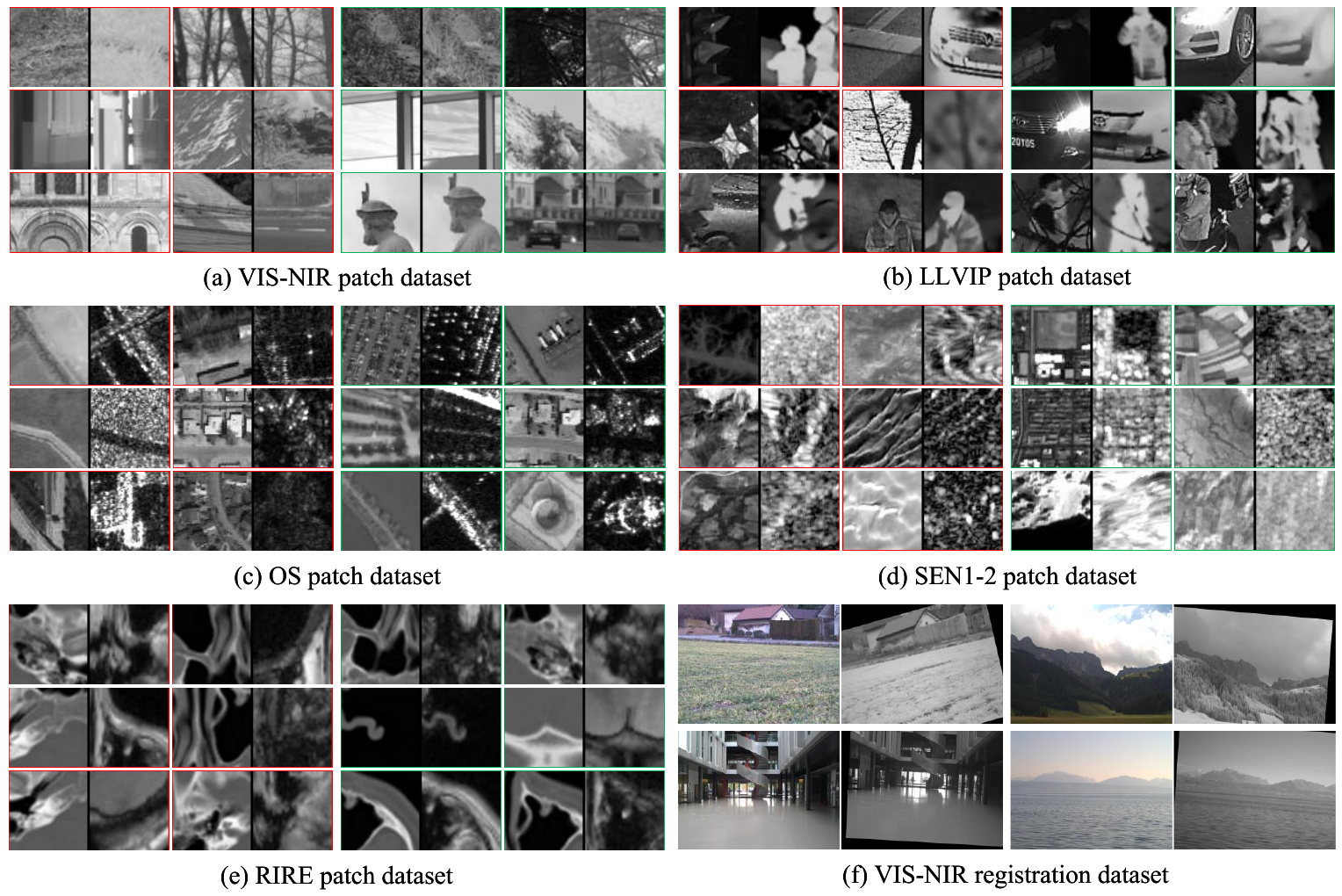}
	\vspace{-18pt}
	\caption{\hlfix{Partial sample display of multiple datasets. The} \textcolor{darkred}{\textit{\hlfix{red}}} \hlfix{boxes denote non-matching sample pairs, and the} \textcolor{darkgreen}{\textit{\hlfix{green}}} \hlfix{boxes denote matching sample pairs.}}
	\label{fig:dataset-intro}
	\vspace{-5pt}
\end{figure*}

\subsection{MLPP Optimization Strategy}
We propose a MLPP optimization strategy that can greatly improve matching performance. The overall loss consists of two parts: reconstruction loss and metric loss. It can be formulated as follows:
\begin{equation}
	{L_{all}} = {L_m} + \alpha \,\sum\limits_{j = 1}^2 {L_{_r}^j} 
\end{equation}
where $L_{all}$, $L_m$, and $L_r^j$ denote the total loss, metric loss, and reconstruction loss, respectively. $\alpha$ is set to 1.

To fully mine the intrinsic features of individual image patches, we constrain the reconstructed image by a combination of perceptual loss~\cite{johnson2016perceptual} and MSE loss. The reconstruction loss is formulated as:
\begin{equation}
	L_r^j = \sqrt {{{({\varphi _p}(P_r^j) - {\varphi _p}({P^j}))}^2}}  + \beta \left( {\sqrt {{{(P_r^j - {P^j})}^2}} } \right)
\end{equation}
where $P_r^j$ and $P^j$ denote the reconstructed image and original image, respectively. $\varphi_p$ denotes a VGG19 network~\cite{simonyan2014very} pretrained on ImageNet~\cite{deng2009imagenet}. $\beta$ is set to 0.1.

\hlfix{To fully capture the feature relations between image patches,} we use multiple metric networks to impose strong constraints on the extracted relation features. On the one hand, we use a metric network in the FIL module to constrain the extracted common features, which is called module loss. On the other hand, we use a metric network on the aggregated private features output at each stage, which is called stage loss. The metric loss is formulated as:
\begin{gather}
	{L_m}(\hat{y}^i,\hat{y}^j,y) = \sum\limits_{i = a}^5 {{L_{module}}(\hat{y}^i,y) + } \sum\limits_{j = b}^5 {{L_{stage}}(\hat{y}^j,y)} \\
	L(\hat{y} ,y) =  - \left[ {y\log (\hat{y}) + (1 - y)\log (1 - \hat{y})} \right]
\end{gather}
where $L_{module}$ and $L_{stage}$ denote module loss and stage loss, respectively. $\hat{y}^i$ and $\hat{y}^j$ are the actual outputs of the FIL modules and aggregated features through the metric network. $y$ denotes the true label. We call the metric network used to output the final similarity the master metric network. Other metric networks are called sub-metric networks. Considering that excessive constraints on low-level feature maps will cause them to focus too much on semantic features and neglect to extract detailed features that should be provided to the main network, we appropriately reduce the strong constraints on low-level feature maps. The experiments in \cref{sec:ablation-study} verify that both $a$ and $b$ should be set to 3.

If the multi-loss strategy is used directly, it will bloat the final model and affect the inference speed. Therefore, we propose a post-pruning strategy. Specifically, before saving the generated model, we prune all branches that do not interfere with the normal output during the inference phase. By using post-pruning, the parameters of the final saved model will be greatly reduced, and the inference speed will be greatly improved. A detailed  experimental verification can be found in \cref{sec:ablation-study}.

\begin{table*}[tb]
	\caption{Comparison of FPR95 among RRL-Net and eighteen methods on the VIS-NIR patch dataset. \textbf{Bold} denotes optimal, \underline{underlined} denotes suboptimal.}
	\label{tab:tab03}
	\centering
	\renewcommand{\arraystretch}{1.0} 
	\resizebox{1.0\textwidth}{!}{
    \setlength{\tabcolsep}{3.5mm}{
		\begin{tabular}{*{10}{c}}
			\toprule
			Models &  Field  & Forest & Indoor & Mountain & Oldbuilding & Street & Urban & Water & Mean    \\
			\midrule
			\multicolumn{10}{c}{Traditional methods}                                                                                                                       \\
			SIFT\textsuperscript{\cite{lowe2004distinctive}[IJCV’04]}  &39.44 &11.39   &10.13    &28.63    &19.69    &31.14   &10.85   &40.33   &23.95   \\
			GISIFT\textsuperscript{\cite{firmenichy2011multispectral}[ICIP’11]}       & 34.75  & 16.63  & 10.63   & 19.52   & 12.54   & 21.80  & 7.21   & 25.78  & 18.60   \\
			LGHD\textsuperscript{\cite{aguilera2015lghd}[ICIP’15]}         & 16.52  & 3.78   & 7.91    & 10.66   & 7.91    & 6.55   & 7.21   & 12.76  & 9.16    \\
			\midrule
			\multicolumn{10}{c}{Descriptor learning}                                                                                                                               \\
			Q-Net\textsuperscript{\cite{aguilera2017cross}[Sensors’17]}        & 17.01  & 2.70   & 6.16    & 9.61     & 4.61   & 3.99   & 2.83    & 8.44   & 6.91   \\
			L2-Net\textsuperscript{\cite{tian2017l2}[CVPR’17]}       & 13.67  & 2.48   & 4.63    & 8.87     & 4.12  & 5.58   & 1.54    & 6.55   & 5.93   \\
			HardNet\textsuperscript{\cite{mishchuk2017working}[NeurIPS’17]}      & 5.61   & 0.15   & 1.50    & 3.14     & 1.10   & 1.93   & 0.69    & 2.29   & 2.05   \\
			SOSNet\textsuperscript{\cite{tian2019sosnet}[CVPR’19]}       & 5.94   & 0.13   & 1.53   & 2.45     & 0.99   & 2.04   & 0.78    & 1.90   & 1.97   \\
			HyNet\textsuperscript{\cite{tian2020hynet}[NeurIPS’20]}        & 4.50   & 0.07   & 1.09    & 1.80     & 0.83   & 0.52   & 0.53    & 1.91   & 1.41  \\
			\midrule
			\multicolumn{10}{c}{Metric learning}                                                                                                                               \\
			Siamese\textsuperscript{\cite{aguilera2016learning}[CVPRW’16]}      & 15.79  & 10.76  & 11.60   & 11.15    & 5.27   & 7.51   & 4.60    & 10.21  & 9.61   \\
			Pseudo-Siamese\textsuperscript{\cite{aguilera2016learning}[CVPRW’16]} & 17.01 & 9.82  & 11.17   & 11.86    & 6.75   & 8.25   & 5.65    & 12.04  & 10.31         \\
			2-channel\textsuperscript{\cite{aguilera2016learning}[CVPRW’16]}    & 9.96   & 0.12   & 4.40    & 8.89     & 2.30   & 2.18   & 1.58    & 6.40  & 4.47     \\
			SCFDM\textsuperscript{\cite{quan2018cross}[ACCV’18]}        & 7.91   & 0.87   & 3.93    & 5.07     & 2.27   & 2.22   & 0.85   & 4.75   & 3.48     \\
			MR\_3A\textsuperscript{\cite{quan2021multi} [TIP’21]}       & 4.21   & 0.11   & 1.12    & 0.87     & 0.67   & 0.56   & 0.43    & 1.90   & 1.23     \\
			AFD-Net\textsuperscript{\cite{quan2019afd,quan2021element}[ICCV’19,TNNLS’21]}  & 3.47   & 0.08   & 1.48    & \underline{0.68}& 0.71  & 0.42   & 0.29    & \underline{1.48} & 1.08  \\
			SPIMNet\textsuperscript{\cite{ko2023spectral}[Inf. Fusion’23]}      & \underline{2.28} & 0.09 & 1.62  & 0.88      & 0.69  & 0.29   & 0.42    & 2.26   & 1.07      \\
			EFR-Net\textsuperscript{\cite{yu2023efficient}[TGRS’23]}      & 2.84   & 0.07   & 1.09    & 0.79     & 0.61   & 0.50   & 0.34    & 1.65   & 0.99  \\
			MFD-Net\textsuperscript{\cite{yu2022multibranch}[TGRS’22]}    & 2.59   & \textbf{0.02} & 1.24 & 0.95  & \underline{0.48} & \textbf{0.24} & \underline{0.12} & \textbf{1.44} & 0.88 \\
			FIL-Net\textsuperscript{\cite{yu2023feature}[TIP’23]}      & 2.61   & \underline{0.04} & \underline{0.98} & 0.73  & 0.49  & 0.34 & \textbf{0.10} & 1.49  & \underline{0.85}    \\
			\textbf{RRL-Net (Ours)}      & \textbf{2.16} & \underline{0.04}    & \textbf{0.86} & \textbf{0.59} & \textbf{0.41} & \underline{0.26} & 0.14 & 1.65 & \textbf{0.76} \\
			\bottomrule
		\end{tabular}
	}}
	\vspace{-0pt}
\end{table*}

\begin{figure*}[tb]
	\centering
	\includegraphics[width=\textwidth]{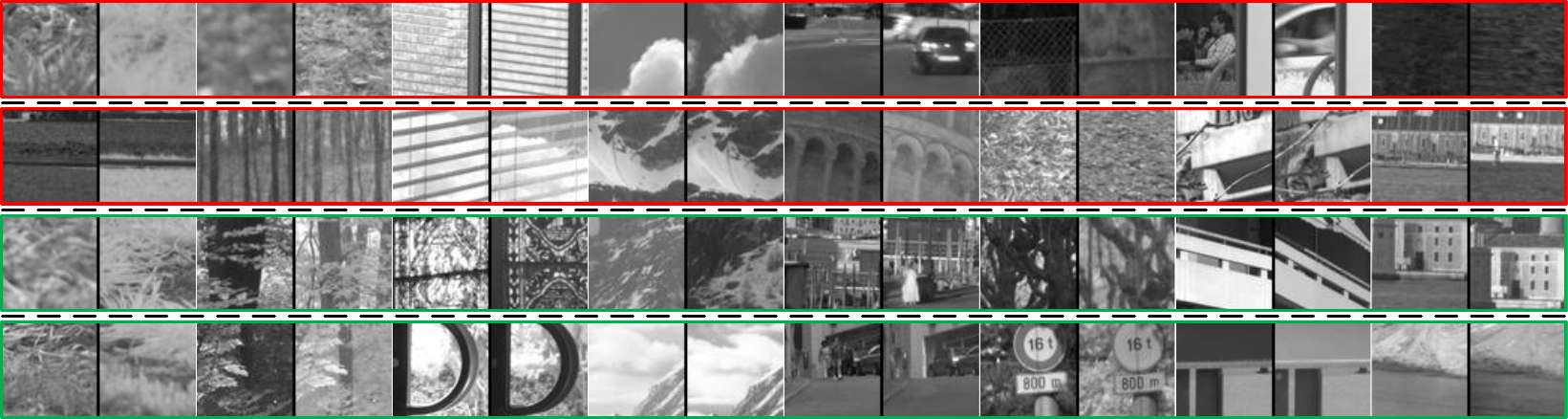}
	\vspace{-15pt}
	\caption{Partial matching results of RRL-Net on the VIS-NIR patch dataset. From top to bottom, the four lines of samples are \textit{misjudged as non-matching}, \textit{misjudged as matching}, \textit{correctly judged as non-matching}, and \textit{correctly judged as matching}.}
	\label{fig:vis-nir-sample-display}
	\vspace{-6pt}
\end{figure*}

\section{Experiment}
\label{experiment}
\subsection{Dataset}
From \cref{fig:dataset-intro}, we experiment on \hlfix{six} different cross-spectral datasets, in which the image patch size is 64×64 pixels. The details are as follows:

\textbf{\textit{VIS-NIR patch dataset~\cite{aguilera2016learning}.}} It is a visible spectrum (VIS) and near-infrared (NIR) image patch matching dataset based on the VIS-NIR scene dataset~\cite{brown2011multi}. Specifically, the dataset contains more than 1.6 million pairs of cross-spectral image patches with equal positive and negative samples, divided into 9 subsets, \hlfix{namely ``Country'', ``Field'', ``Forest'', ``Indoor'', ``Mountain'', ``Oldbuilding'', ``Street'', ``Urban'', and ``Water''.} For the construction of this dataset, VIS image patches cropped based on SIFT keypoints as the center are first acquired. Secondly, half of the VIS image patches and NIR image patches whose keypoint coordinates are consistent form matching samples. Finally, the other half of the VIS image patches and random NIR image patches form non-matching samples. Consistent with~\cite{quan2021multi,yu2023feature,aguilera2016learning,quan2019afd,quan2021element,yu2022multibranch,yu2023efficient,en2018ts,aguilera2017cross,quan2018cross,ko2023spectral}, only the ``Country'' subset is used for training, and the other eight subsets are used for testing.

\hlfix{\textbf{\textit{LLVIP patch dataset.}} It is a VIS and long-wave infrared (LWIR) image patch matching dataset which we built based on the LLVIP dataset~\cite{jia2021llvip}. To construct the dataset, we first apply SIFT~\cite{lowe2004distinctive} to extract keypoints from pre-registered image pairs and remove those located near the image boundaries that could not yield valid patches. Then, matching point pairs are constructed by comparing the coordinate consistency of the keypoints in the paired images, and non-matching point pairs are randomly sampled according to the number of matching point pairs. The image patches are cropped to 64×64 pixels with the keypoints as the center. Finally, the training set (197,408 pairs) and the test set (10,528 pairs) are generated according to the original sample division in the LLVIP dataset. This dataset covers a variety of complex lighting conditions (mainly low-light scenes), and there are significant nonlinear differences between VIS and LWIR images. \cref{fig:dataset-intro} shows some samples from the LLVIP patch dataset.}

\textbf{\textit{OS patch dataset~\cite{yu2023efficient}.}} It is a VIS and synthetic aperture radar (SAR) image patch matching dataset based on the OS dataset~\cite{xiang2020automatic}. Each image patch is cropped at the center of the keypoints obtained by the SIFT method. Pairs of cross-spectral image patches with the same keypoints are identified as matching samples. Pairs of cross-spectral image patches with different keypoints are identified as non-matching samples. The number of generated non-matching samples is the same as the number of matching samples. The dataset contains a total of 123,676 samples, of which the training set and the test set contain 98,940 and 24,736 samples. From \cref{fig:dataset-intro}, compared with the VIS-NIR patch dataset, the sample pairs in the OS patch dataset have more obvious nonlinear differences.

\textbf{\textit{SEN1-2 patch dataset~\cite{yu2023feature}.}} It is a VIS and SAR image patch matching dataset based on the SEN1-2 dataset~\cite{schmitt2019sen12ms}. The construction process of this dataset is consistent with the OS patch dataset. This dataset has more samples, containing a total of 800,000 samples, of which the training set and the test set contain 600,000 and 200,000 respectively. From \cref{fig:dataset-intro}, compared to the OS patch dataset, the scenes in the SEN1-2 patch dataset have less semantic feature information.

\begin{figure}[!t]
\begin{minipage}{\columnwidth}
	\centering
	\includegraphics[width=\columnwidth]{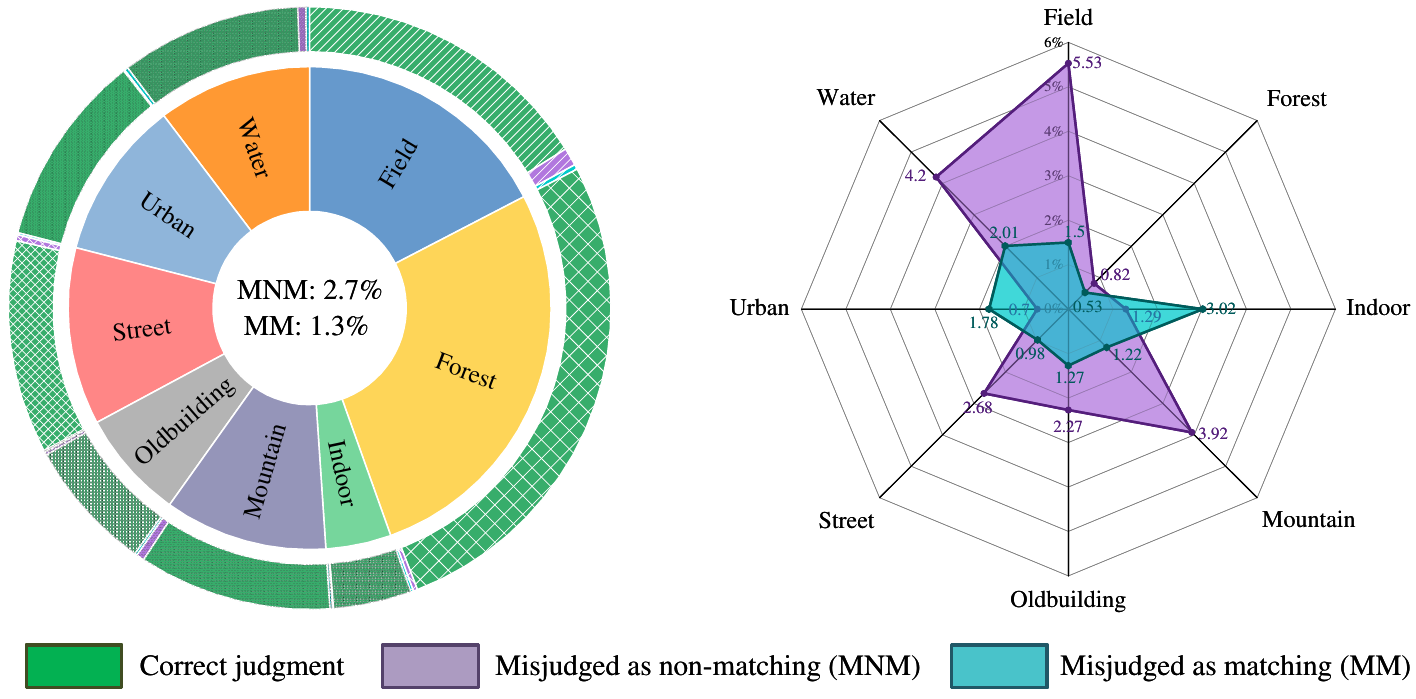}
	\vspace{-17pt}
	\caption{The misjudgment percentages of RRL-Net on eight subsets.}
	\label{fig:vis-nir-misjudgment-percentages}
	\vspace{-10pt}
\end{minipage}

\vspace{19pt}

\begin{minipage}{\columnwidth}
	\centering
	\includegraphics[width=\columnwidth]{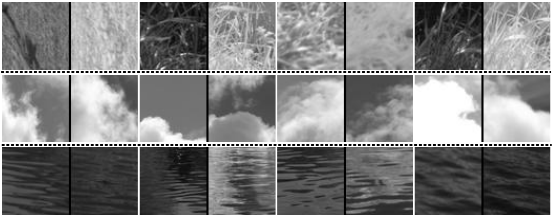}
	\vspace{-17pt}
	\caption{\hlfix{Some samples falsely identified as non-matching due to environmental influences. From top to bottom, each row comes from the ``field'', ``mountain'', and ``water'' subsets respectively.}}
	\label{fig:vis-nir-grass-water-misjudment}
	\vspace{-10pt}
\end{minipage}

\vspace{19pt}

\begin{minipage}{\columnwidth}
	\centering
	\includegraphics[width=\columnwidth]{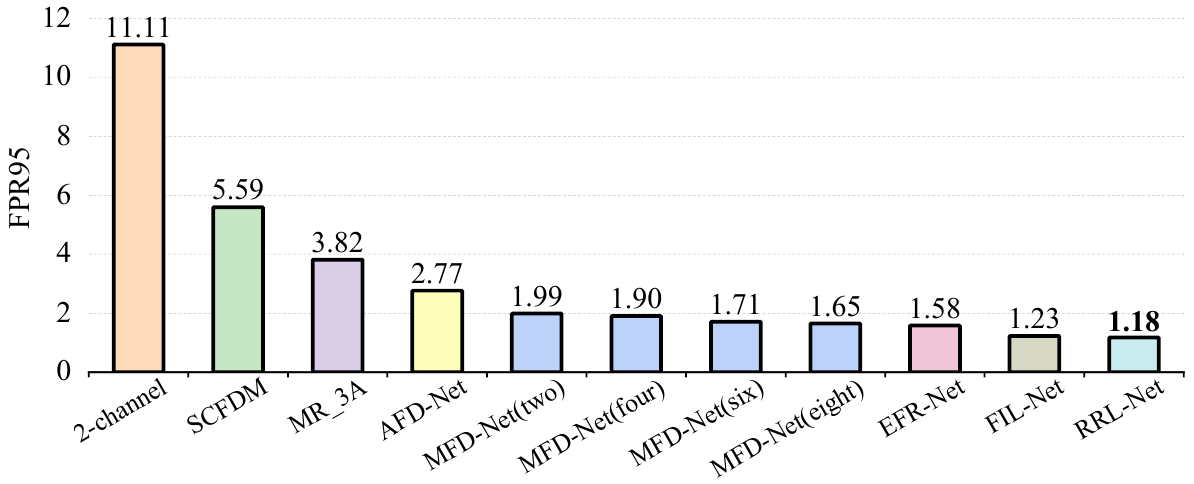}
	\vspace{-18pt}
	\caption{\hlfix{FPR95 comparison of various excellent image patch matching methods on the LLVIP patch dataset.}}
	\label{fig:vis-lwir-methods-fpr95}
	\vspace{-10pt}
\end{minipage}

\vspace{19pt}

\begin{minipage}{\columnwidth}
	\centering
	\includegraphics[width=\columnwidth]{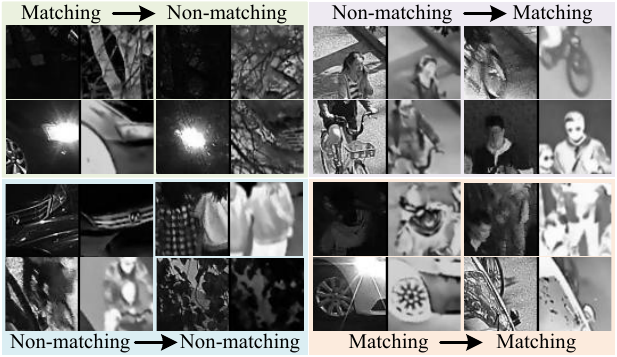}
	\vspace{-17pt}
	\caption{\hlfix{Partial results of RRL-Net on the LLVIP patch dataset. The left and right sides of the arrow denote the true labels and predicted results.}}
	\label{fig:vis-lwir-sample-display}
	\vspace{-10pt}
\end{minipage}
\vspace{-5pt}
\end{figure}

\hlfix{\textbf{\textit{RIRE patch dataset.}} It is a computed tomography (CT) and magnetic resonance imaging (MRI) image patch matching dataset which we built based on the public RIRE (Retrospective Image Registration Evaluation) dataset. MRI includes three imaging types (T1, T2, PD), so each CT sample corresponds to three image pairs: CT-MRI (T1), CT-MRI (T2), CT-MRI (PD). For brevity, we collectively refer to them as CT-MRI. For dataset production, except that it uses the FAST~\cite{rosten2006machine} to extract keypoints, the remaining operations are consistent with the LLVIP patch dataset production. Finally, the generated RIRE patch dataset contains a total of 77,194 pairs, including 61,755 pairs of training samples and 15,439 pairs of test samples. The image patch size is 64 × 64 pixels. \cref{fig:dataset-intro} shows some samples from the RIRE patch dataset.}

\textbf{\textit{VIS-NIR registration dataset.}} This registration dataset is a VIS and NIR registration dataset that we construct based on the VIS-NIR scene dataset~\cite{brown2011multi}. It has 160 test sample pairs, which come from the first 20 pairs of each subset. The image pair to be registered is generated by performing an affine transformation including rotation, translation and scaling on the NIR images in the original registered image pair. Consistent with the experiments on the VIS-NIR patch dataset, the experiments are trained on the ``Country'' subset and tested on the other eight subsets. The dataset we made can be available at \url{https://github.com/YuChuang1205/VIS-NIR-registration-dataset}

\begin{figure*}[t]
	\centering
	\includegraphics[width=\textwidth]{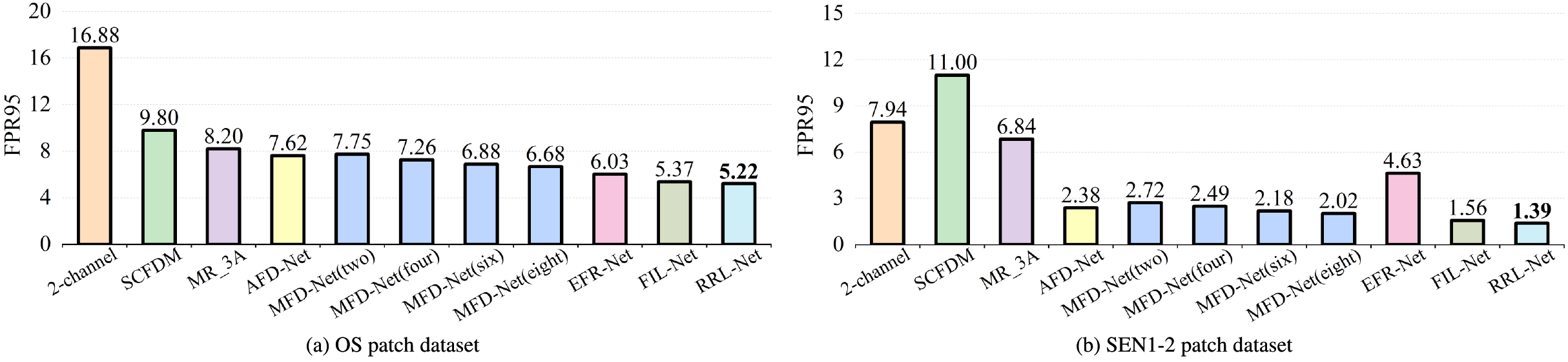}
	\vspace{-16pt}
	\caption{Performance comparison on two VIS-SAR image patch  matching datasets.}
	\label{fig:vis-sar-methods-fpr95}
	\vspace{-0pt}
\end{figure*}

\begin{figure*}[t]
	\centering
	\includegraphics[width=\textwidth]{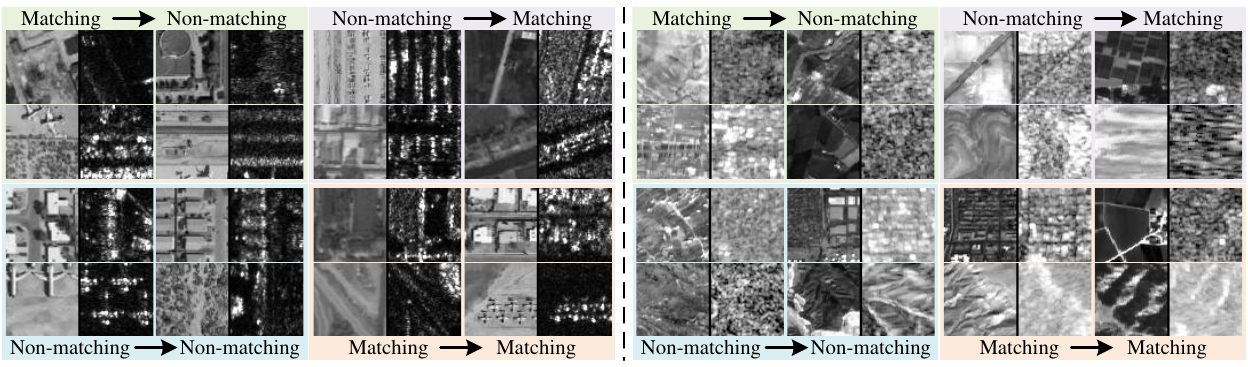}
	\vspace{-16pt}
	\caption{Partial results of RRL-Net on two VIS-SAR image patch matching datasets. The left and right of the dotted line are the OS patch dataset and the SEN1-2 patch dataset respectively. The left and right sides of the arrow denote the true labels and predicted results.}
	\label{fig:vis-sar-sample-display}
	\vspace{-3pt}
\end{figure*}

\subsection{Experimental Settings}
\textbf{\textit{Training details.}} The epochs, learning rate, and batch size are set to 40, $2e^{-4}$, and 64, respectively. \hlfix{The GPU is RTX 2080Ti 11G and the CPU is Intel i7-9700K CPU @ 3.60GHz. The training time of the proposed RRL-Net on the VIS-NIR, LLVIP, OS, and SEN1-2 patch datasets is 19.3 hours, 13.9 hours, 6.9 hours, and 40.7 hours, respectively. At the same time, to} avoid the network from falling into overfitting, data augmentation operations such as flipping, rotation, and contrast enhancement are used to generate lots of images and improve model generalization. \hlfix{In addition, consistent with previous studies~\cite{quan2021multi,yu2023feature,quan2021element,yu2022multibranch,yu2023efficient}, RRL-Net is trained from scratch. To ensure reproducibility, we release all source code and step-by-step instructions at the} \href{https://github.com/YuChuang1205/RRL-Net}{\hlfix{project link}} \hlfix{in the Abstract.}

\textbf{\textit{Evaluation Metrics.}} For image patch matching, the false positive rate at 95\% recall (FPR95)~\cite{quan2021multi,yu2023feature,aguilera2016learning,quan2019afd,quan2021element,yu2022multibranch,yu2023efficient,en2018ts,aguilera2017cross,quan2018cross,ko2023spectral} and area under the curve (AUC) are used. For cross-spectral image registration, the root mean square error (RMSE) and image registration rate $I_{rr}$ are used~\cite{quan2022self,quan2022deep,yu2023feature}. An image after affine transformation with a RMSE less than 5 is called a registered image. In the results presented, \hlfix{\textbf{bold} denotes the optimal results and \underline{underline} denotes the suboptimal results. The MFD-Net defaults to the eight-branch structure.}

\subsection{Comparison with State-of-the-Art Methods} 
\textbf{\textit{1) VIS-NIR image patch matching.}} We conduct experiments with 18 state-of-the-art methods on the VIS-NIR patch dataset. \hlfix{From the individual results of the eight subsets in \cref{tab:tab03},} the proposed RRL-Net achieved optimal results on four subsets, and the results of the remaining four subsets are not much different from the optimal results. \hlfix{At the same time, from the mean results of eight subsets in \cref{tab:tab03}, RRL-Net achieves SOTA result. Specifically, compared with HyNet,} the descriptor learning method with the best results, our RRL-Net reduces the mean FPR95 by 46.1\% (from 1.41 to 0.76). Compared with the recent metric learning methods SPIMNet, EFR-Net, MFD-Net and FIL-Net, RRL-Net reduced mean FPR95 by 29.0\% (from 1.07 to 0.76), 23.2\% (from 0.99 to 0.76), 13.6\% (from 0.88 to 0.76) and 10.6\% (from 0.85 to 0.76). Notably, compared to MFD-Net in 2022, FIL-Net, which has the best results in 2023, only decreases by 3.4\% (from 0.88 to 0.85). \hlfix{In addition, according to the statistics, compared with FIL-Net, RRL-Net reduces the number of parameters by 72.7\% (from 0.33 to 0.09) and the single-sample inference time by 43.7\% (from 0.71×$10^{-3}$ to 0.40 ×$10^{-3}$).} \hlB{Notably, the ``Country'' subset of the VIS-NIR patch dataset used for training contains only 52 original training image pairs. RRL-Net achieves excellent results on the remaining 8 subsets, demonstrating its effectiveness even under limited data conditions.}

To qualitatively analyze the matching effect of RRL-Net in the VIS-NIR scene, \cref{fig:vis-nir-sample-display} shows some matching results on the VIS-NIR patch dataset. From the samples misjudged as non-matching, most individual samples in pairs have unclear features. At the same time, due to the movement of objects, there are samples that are marked as matching but visually non-matching. From the samples misjudged as matching, most samples have obviously similar semantic features. From the samples that are correctly judged as non-matching, RRL-Net can correctly distinguish non-matching samples with partially similar semantic features. From the samples that are correctly judged as matching, there are nonlinear differences between cross-spectral image patches. \hlfix{These prove} that RRL-Net is robust for VIS-NIR scenarios.

\begin{table*}[t]
	\begin{minipage}[t]{0.51\textwidth}
		\centering
		\vspace{0pt}
		\begin{subtable}{\textwidth}
			\renewcommand{\arraystretch}{1} 
			\centering
			\renewcommand{\arraystretch}{1.11}
			\resizebox{\textwidth}{!}{
				
				\begin{tabular}{c|ccc|c}
					\toprule
					& \multicolumn{3}{c|}{Variants}         &                          \\
					\multirow{-2}{*}{\quad Methods \quad} &\quad MGLA \quad  &\quad FIL \quad &\quad MLPP \quad & \multirow{-2}{*}{\quad FPR95 ↓ \quad}   \\
					\midrule
					\quad Our-w/o MGLA \quad  & \quad \ding{55} \quad &\quad \ding{51}\quad &\quad \ding{51} \quad &\quad 1.14 (-33.3\%) \quad \\
					\quad Our-w/o FIL \quad  &\quad \ding{51} \quad & \quad \ding{55} \quad &\quad \ding{51} \quad  &\quad 1.02 (-25.5\%)  \quad \\
					\quad Our-w/o MLPP \quad  &\quad \ding{51} \quad & \quad \ding{51} \quad & \quad \ding{55} \quad  &\quad 1.01 (-24.8\%) \quad \\
					\midrule
					\rowcolor[HTML]{E0E0E0} 
					\textbf{RRL-Net (Ours)} &\quad \ding{51} \quad & \quad \ding{51} \quad & \quad \ding{51} \quad  &  \textbf{\quad 0.76 \quad}  \\
					\bottomrule      
			\end{tabular}}
			\caption{Break-down ablation}
			\label{tab:Ablation-table-break-down}
		\end{subtable}
		
		\vspace{0.52em} 
		\begin{subtable}{\textwidth}
			\vspace{-0.1\baselineskip} 
                 \renewcommand{\arraystretch}{1.15}
			\resizebox{\textwidth}{!}{
				\begin{tabular}{c|ccc|c|c|c}
					\toprule
					& \multicolumn{3}{c|}{CL}        &     &           &     \\
					\multirow{-2}{*}{Methods} & 1   & 2    & 3    & \multirow{-2}{*}{CC} & \multirow{-2}{*}{FPR95 ↓} & \multirow{-2}{*}{Parameters↓} \\
					\midrule
					Variant 1   & \ding{55}   & \ding{55}   & \ding{51}   & \ding{51}   & 0.83 (-8.4\%)  & 12,648,441 (-25.7\%) \\
					Variant 2   & \ding{55}    & \ding{55}  & \ding{51}  & \ding{55}     & 0.91 (-16.5\%) & 12,556,601 (-25.1\%) \\
					Variant 3    & \ding{55}   & \ding{51}   & \ding{55}   & \ding{51}   & 0.80 (-5.0\%)  & 11,072,153 (-15.1\%) \\
					Variant 4   & \ding{55}   & \ding{51}   & \ding{55}   & \ding{55}    & 0.84 (-9.5\%)  & 10,980,313 (-14.4\%) \\
					Variant 5    & \ding{51}  & \ding{55}  & \ding{55}   & \ding{51}  & 0.82 (-7.3\%)  & 9,495,865 (-1.0\%)   \\
					\midrule
					\rowcolor[HTML]{E0E0E0} 
					\textbf{ALFE (Ours)} & \ding{51}   & \ding{55}  & \ding{55}  & \ding{55} & \textbf{0.76} & \textbf{9,404,025} \\ 
					\bottomrule
			\end{tabular}}
			\caption{Exploration of ALFE network}
			\label{tab:Ablation-table-ALFE}
		\end{subtable}
	\end{minipage}
	\hfill
	\begin{minipage}[t]{0.49\textwidth}
		\centering
		\vspace{0pt}
		\begin{subtable}{1\textwidth}
			\renewcommand{\arraystretch}{1.11} 
			\resizebox{1\textwidth}{!}{
				\setlength{\tabcolsep}{2.68mm}{
					
					\begin{tabular}{c|c|c}
						\toprule
						
						\raisebox{0.8ex}{\rule{0pt}{13pt}  Attention modules} & \raisebox{0.8ex}{\rule{0pt}{13pt} FPR95 $\downarrow$} & \raisebox{0.8ex}{\rule{0pt}{13pt} Parameters $\downarrow$} \\
						\midrule
						Our-w/o MGLA                 & 1.14 (-33.3\%)         & \textbf{9,403,905 (+120)}         \\
						SE (r=4)\textsuperscript{\cite{hu2018squeeze}}     & 0.93 (-18.3\%)         & 9,492,441 (-0.9\%)                \\
						SE (r=8)\textsuperscript{\cite{hu2018squeeze}}     & 0.98 (-22.4\%)         & 9,448,669 (-0.5\%)                \\
						SE (r=16)\textsuperscript{\cite{hu2018squeeze}}    & 1.07 (-29.0\%)         & 9,426,783 (-0.2\%)                \\
						ECA\textsuperscript{\cite{wang2020eca}}          & 0.83 (-8.4\%)          & 9,403,953 (+72)                   \\
						ELA\textsuperscript{\cite{yu2023efficient}}          & 0.81 (-6.2\%)          & 9,404,025                         \\
						CBAM (r=4)\textsuperscript{\cite{woo2018cbam}}   & 0.97 (-21.6\%)         & 9,493,421 (-0.9\%)                \\
						SNL\textsuperscript{\cite{cao2019gcnet}}          & 0.94 (-19.1\%)         & 9,580,491 (-1.8\%)                \\
						GC (r=4)\textsuperscript{\cite{cao2019gcnet}}     & 0.90 (-15.6\%)         & 9,494,435 (-1.0\%)                \\
						HGLA (Ours)                              & 1.08 (-29.6\%)         & 9,403,941 (+84)                   \\
						WGLA (Ours)                              & 1.05 (-27.6\%)         & 9,403,941 (+84)                   \\
						CGLA (Ours)                              & 0.79 (-3.8\%)          & 9,403,953 (+72)                   \\
						\midrule
						\rowcolor[HTML]{E0E0E0} 
						\textbf{MGLA (Ours)} & \cellcolor[HTML]{E0E0E0}\textbf{0.76} & \cellcolor[HTML]{E0E0E0}9,404,025   \\
						\bottomrule
					\end{tabular}
					
			} }
			\caption{Exploration of MGLA module}
			\label{tab:Ablation-table-MGLA}
			
		\end{subtable}
	\end{minipage}
       \vspace{-5pt}
	\caption{Ablation experiments on VIS-NIR patch dataset. \textbf{CL}: Convolutional layers before each MGLA module, \textbf{CC}: Cross-layer connection. $r$ is compression ratio. HGLA, WGLA, and CGLA are single dimension $H$, $W$, and $C$ global-to-local attention.}
	\vspace{-5pt}
\end{table*}

To further quantitatively analyze the misjudgment of matching samples and non-matching samples by RRL-Net on the VIS-NIR patch dataset, we present the misjudgment percentage of eight subsets, as shown in \cref{fig:vis-nir-misjudgment-percentages}. On the whole, the percentage of misjudged as matching and misjudged as non-matching is very small, which fully demonstrates the high discriminability of our RRL-Net. In addition, we find that the percentage of misjudged as non-matching is greater than the percentage of misjudged as matching. The reason is that the experiment is trained on ``Country'' and tested on the other eight subsets. Therefore, the matching network can only approximately learn the domain information in the VIS-NIR scenario. There are still different nonlinear differences in the other subsets. At the same time, it is also due to the spatiotemporal difference of collected images. Some moving objects or shooting position deviations will cause the samples that are finally marked as matching to appear visually non-matching. \hlfix{From the results of each subset, the percentage of samples in ``Field'', ``Mountain'' and ``Water'' that are misjudged as non-matching is significantly larger. On the one hand, it is due to the influence of the environment. From \cref{fig:vis-nir-grass-water-misjudment}, wind has serious spatiotemporal effects on the targets (grass, clouds, water waves) in the corresponding scene. On the other hand, this is due to the influence of the imaging conditions.} The scenes in the three subsets are more likely to be too bright or too dark, resulting in a scarcity of single image features in the image patch pairs. This leads to significant nonlinear differences between image patches. \hlB{In addition to the above direct reasons, from the perspective of network structure composition, it is due to the local dependency modeling of the MGLA module and the differential amplification effect of the FIL module. Specifically, on the one hand, the MGLA module emphasizes local dependencies while extracting global features. In most scenarios, this mechanism can significantly improve the accuracy of cross-spectral matching. However, in highly dynamic environments such as grass, clouds, or water ripples, spatiotemporal inconsistencies can cause local regions to contain noise features that change rapidly over time. These instantaneous differences can easily be misjudged by the model as discriminative features, thereby weakening the stabilizing effect brought by cross-spectral global consistency. On the other hand, the FIL module aims to highlight the differences in private features between cross-spectral image patches and improve the discrimination ability by mining deep feature relationships. This design can significantly improve the matching effect in most scenarios. However, when the input contains too much dynamic noise, these unstable components may be over-amplified, ultimately causing the discrimination results to deviate from the true semantic correspondence.}

\begin{figure*}[tb]
	\centering
	\includegraphics[width=\textwidth]{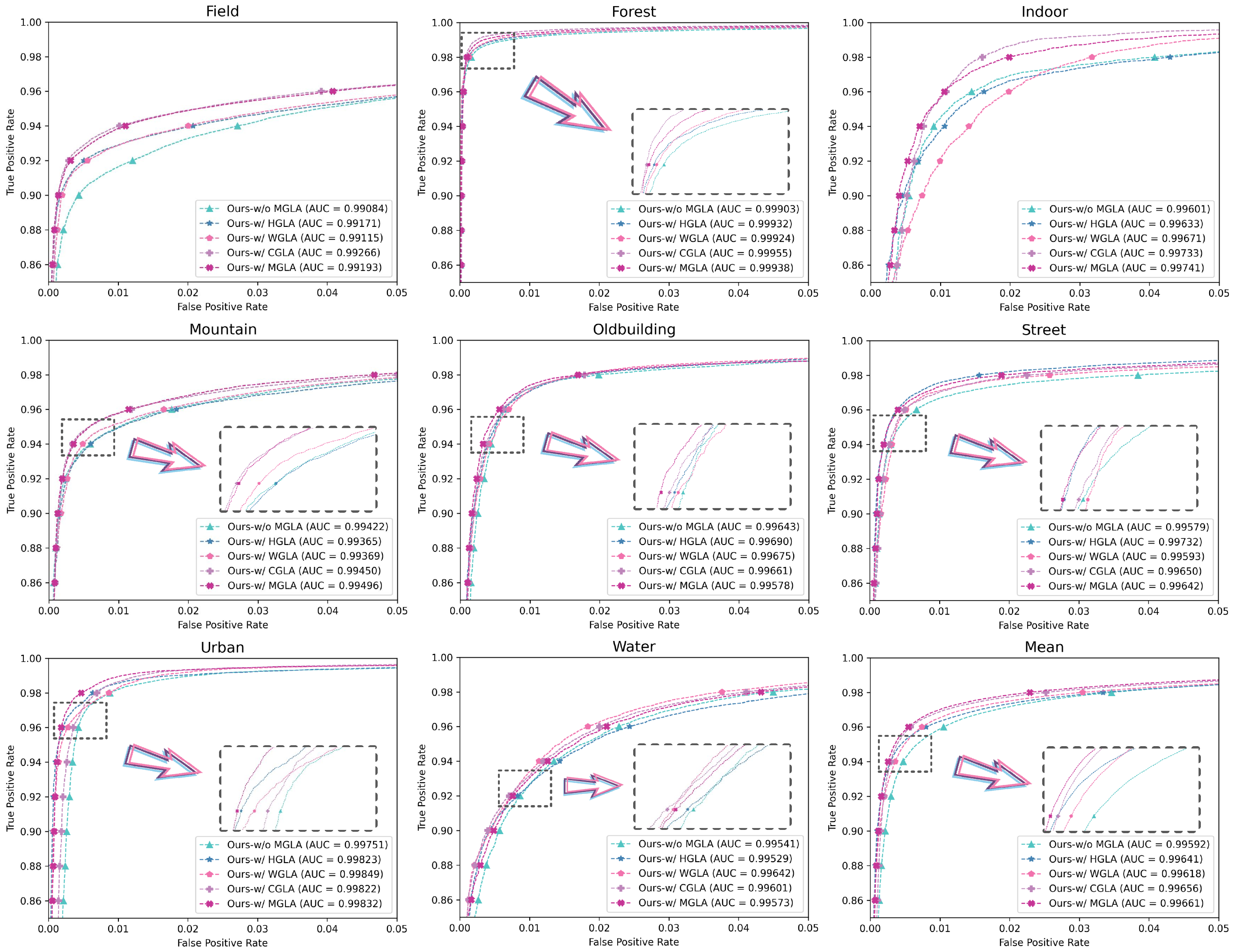}
	\vspace{-15pt}
	\caption{\hlfix{Ablation experiments of the MGLA module on the VIS-NIR patch dataset. The dotted line is the local enlarged area. The receiver operating characteristic (ROC) curves without using the MGLA module are mostly at the lowest, and those with using the MGLA module are mostly at the highest. From the “Mean” subgraph, compared with not using the MGLA module, the \textit{AUC} is significantly improved whether using a single-dimensional global-to-local module (HGLA, WGLA, CGLA) or a multi-dimensional global-to-local module (MGLA). In addition, using the final MGLA module has the best \textit{AUC}.}}
	\label{fig:roc}
	\vspace{-5pt}
\end{figure*}

\hlfix{\textbf{\textit{2) VIS-LWIR image patch matching.}} To explore the performance of RRL-Net in VIS-LWIR cross-spectral scenarios, we compare it with multiple image patch matching methods on the LLVIP patch dataset. From \cref{fig:vis-lwir-methods-fpr95}, compared with other methods, RRL-Net drops by 4.1\% (from 1.23 to 1.18) - 89.4\% (from 11.11 to 1.18) on FPR95. The results verify the effectiveness of RRL-Net in this cross-spectral scenario.}

\hlfix{To further qualitatively analyze the discriminative ability of RRL-Net on the LLVIP patch dataset, we show the matching results of some samples in \cref{fig:vis-lwir-sample-display}. For the samples that are misjudged as non-matching, most of them are due to severe underexposure of VIS image patches or partial overexposure due to vehicle high beams, resulting in significant information loss. For the samples that are misjudged as matching, most of them are due to keypoints being adjacent but not overlapping, and image patches having strong semantic similarity, which leads to model misjudgment. From the samples that are correctly judged as non-matching, it shows that RRL-Net can effectively distinguish non-matching patch pairs with partial structural similarities. From the samples that are correctly judged as matching, it shows that RRL-Net can achieve accurate alignment of cross-modal semantics even when there are significant nonlinear differences between VIS and LWIR. In addition, according to statistics, the samples that are misjudged as non-matching and misjudged as matching account for only 1.8\% and 0.7\% respectively, indicating that the overall matching ability of the model is effective and stable. In summary, RRL-Net demonstrates excellent cross-spectral image patch matching capabilities for VIS-LWIR scenarios under low illumination and high nonlinear differences conditions.}

\textbf{\textit{3) VIS-SAR image patch matching.}} We compare \hlfix{the proposed RRL-Net} with 7 state-of-the-art methods on the OS patch dataset and SEN1-2 patch dataset. From \cref{fig:vis-sar-methods-fpr95}a, our RRL-Net achieves SOTA results on the OS patch dataset. Compared with the methods AFD-Net, MFD-Net(eight), EFR-Net and FIL-Net, RRL-Net reduced FPR95 by 31.5\% (from 7.62 to 5.22), 21.9\% (from 6.68 to 5.22), 13.4\% (from 6.03 to 5.22) and 2.8\% (from 5.37 to 5.22), respectively. This shows that our RRL-Net also has good robustness for VIS-SAR cross-spectral image patch matching scenarios. From \cref{fig:vis-sar-methods-fpr95}b, first, compared to MatchNet, Siamese and Pseudo-Siamese, the other image patch matching methods have achieve significantly better matching results. This shows that learning the feature relations between image patches for the image patch matching task can extract richer discriminative features, especially for image pairs with large nonlinear differences. Secondly, the results of EFR-Net are significantly worse than AFD-Net, MFD-Net, FIL-Net and RRL-Net. This is because EFR-Net only uses high-level feature maps for feature relation learning, while AFD-Net, MFD-Net, FIL-Net and RRL-Net adopt a multi-stage feature map aggregation structure. For image patch pairs with large nonlinear differences, extracting multi-stage feature maps can significantly improve the extraction of discriminative features. The results of MR\_3A, which also uses only high-level feature maps, also illustrate it. At the same time, combined with the results on the OS patch dataset, the FPR95 of EFR-Net and MR\_3A are not much different between the two datasets, while the FPR95 of AFD-Net, MFD-Net, FIL-Net and RRL-Net are significantly different between the two datasets. On the one hand, it is because the multi-stage feature map aggregation structure requires more samples for network optimization, and the number of samples in the SEN1-2 patch dataset is more than 6 times that of the OS patch dataset. On the other hand, compared to the SEN1-2 patch dataset, the samples in the OS patch dataset have more obvious semantic features, such as houses, airplanes, EFR-Net and MR\_3A focus more on high-level semantic features. Finally, our RRL-Net achieves state-of-the-art result on the SEN1-2 patch dataset. Compared with the image patch matching methods AFD-Net, MFD-Net(eight), EFR-Net and FIL-Net, RRL-Net reduces the FPR95 by 41.6\% (from 2.38 to 1.39), 31.2\% (from 2.02 to 1.39), 70.0\% (from 4.63 to 1.39) and 10.9\% (from 1.56 to 1.39).

To qualitatively analyze the matching effect of RRL-Net in the VIS-SAR scene, we display some matching results on the OS patch dataset and SEN1-2 patch dataset as shown in \cref{fig:vis-sar-sample-display}. From the samples misjudged as non-matching, there is a common problem of feature scarcity in SAR images. For texture or semantic features in VIS images, there are almost no obvious corresponding features in SAR images. From the samples misjudged as matching, There are large area similar semantic features in VIS and SAR images. From the samples that are correctly judged, RRL-Net can also correctly identify non-matching samples with partially similar semantic features and matching samples with pixel-level nonlinear differences. This proves that RRL-Net has excellent robustness for VIS-SAR image patch matching scenarios.

\begin{table*}[t]
\caption{Ablation experiments of MLPP optimization strategy on VIS-NIR patch dataset. \textbf{SN}: Sub-metric network, \textbf{RB}: Reconstruction branches, \textbf{RB-Loss}: Reconstruction branch loss, \textbf{ML}: MSE loss, \textbf{PL}: Perceptual loss, \textbf{PP}: Post-pruning.}
	\vspace{-3pt}
	\label{tab:tab02}
	\centering
	\renewcommand{\arraystretch}{1.1} 
	\resizebox{1.0\textwidth}{!}{
		\setlength{\tabcolsep}{2mm}{
\begin{tabular}{c|ccccc|c|c|c|c}
\hline
                          &                       &                      & \multicolumn{2}{c}{RB Loss} &                      &                           &                                &                                                                                               &                                                                                             \\
\multirow{-2}{*}{Methods} & \multirow{-2}{*}{SMN} & \multirow{-2}{*}{RB} & ML           & PL           & \multirow{-2}{*}{PP} & \multirow{-2}{*}{FPR95 ↓} & \multirow{-2}{*}{Parameters ↓} & \multirow{-2}{*}{\begin{tabular}[c]{@{}c@{}}Inference time (s) ↓\\ \hlfix{(GTX 2080Ti)}\end{tabular}} & \multirow{-2}{*}{\begin{tabular}[c]{@{}c@{}}\hlfix{Inference time (s) ↓}\\ \hlfix{(GTX 3090)}\end{tabular}} \\ \hline
Ours-w/o MLPP             & \ding{55}           & \ding{55}           & \ding{55}   & \ding{55}   & \ding{55}           & 1.01 (-24.8\%)            & \textbf{9,404,025}             & \textbf{0.40×10\textsuperscript{-3}}                                                                            & \hlfix{\textbf{0.30×10\textsuperscript{-3}}}                                                                          \\ \hline
Variant 1                 & \ding{55}            & \ding{55}           & \ding{55}   & \ding{55}   & \ding{55}           & 0.95 (-20.0\%)            & 10,190,175 (-7.7\%)            & 0.51×10\textsuperscript{-3} (-21.6\%)                                                                           & \hlfix{0.41×10\textsuperscript{-3} (-26.8\%)}                                                                         \\
Variant 2                 & \ding{51}            & \ding{55}           & \ding{55}   & \ding{55}   & \ding{55}           & 0.83 (-8.4\%)             & 12,043,262 (-21.9\%)           & 0.44×10\textsuperscript{-3} (-9.1\%)                                                                            & \hlfix{0.32×10\textsuperscript{-3} (-6.3\%)}                                                                          \\
Variant 3                 & \ding{51}            & \ding{51}           & \ding{55}   & \ding{51}   & \ding{55}           & 0.79 (-3.8\%)             & 12,829,276 (-26.7\%)           & 0.54×10\textsuperscript{-3} (-25.9\%)                                                                           & \hlfix{0.43×10\textsuperscript{-3} (-30.2\%)}                                                                         \\
Variant 4                 & \ding{51}            & \ding{51}           & \ding{51}   & \ding{55}   & \ding{55}           & 0.79 (-3.8\%)             & 12,829,276 (-26.7\%)           & 0.54×10\textsuperscript{-3} (-25.9\%)                                                                           & \hlfix{0.43×10\textsuperscript{-3} (-30.2\%)}                                                                         \\
Variant 5                 & \ding{51}            & \ding{51}           & \ding{51}   & \ding{51}  & \ding{55}           & \textbf{0.76}             & 12,829,276 (-26.7\%)           & 0.54×10\textsuperscript{-3} (-25.9\%)                                                                           & \hlfix{0.43×10\textsuperscript{-3} (-30.2\%)}                                                                         \\ \hline
\rowcolor[HTML]{E0E0E0} 
\textbf{Ours-w/ MLPP}     & \ding{51}            & \ding{51}           & \ding{51}   & \ding{51}   & \ding{51}           & \textbf{0.76}             & \textbf{9,404,025}             & \textbf{0.40×10\textsuperscript{-3}}                                                                            & \hlfix{\textbf{0.30×10\textsuperscript{-3}}}                                                                          \\ \hline
\end{tabular}
}
}
\end{table*}

\begin{figure*}[tb]
	\centering
	\includegraphics[width=\textwidth]{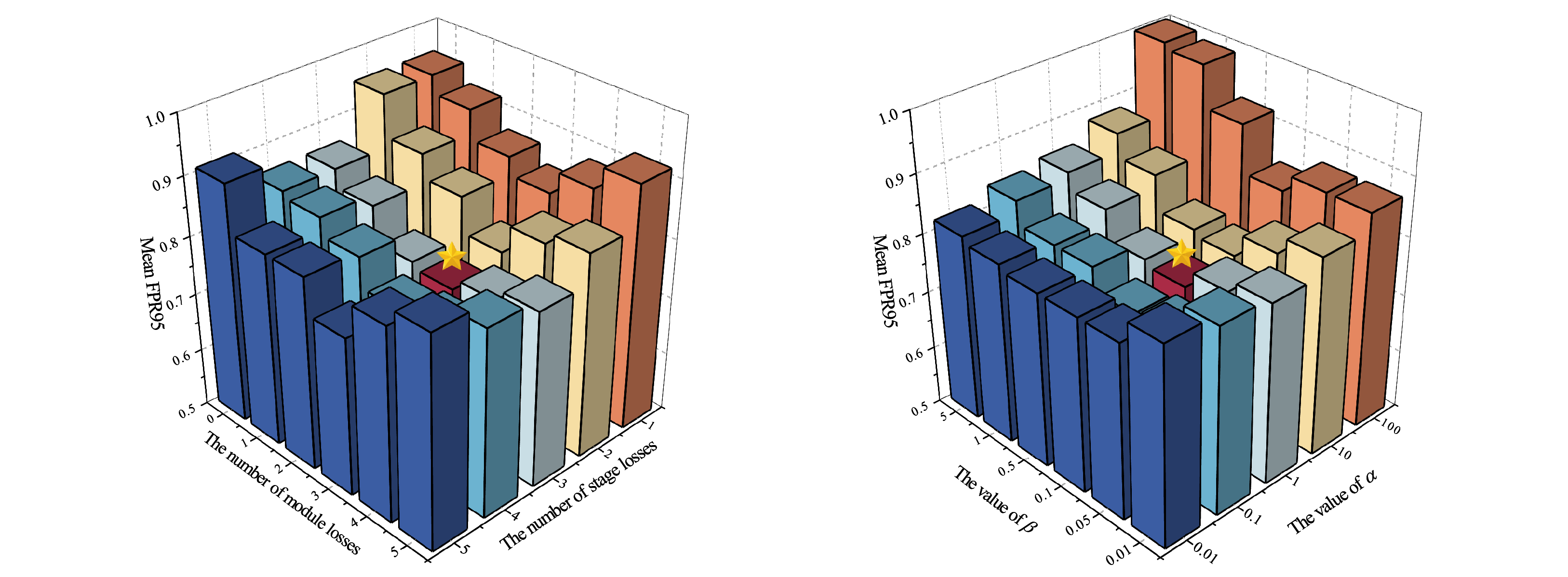}
	\vspace{-12pt}
	\caption{\hlfix{Exploration of composition details in the MLPP optimization strategy on the VIS-NIR patch dataset. On the left are the results of different combinations of stage loss and module loss on the VIS-NIR patch dataset. The loss is added from higher levels to lower levels. On the right are the results of different combinations of $\alpha$ and $\beta$ in overall loss. The} \textcolor{darkred}{\textit{\hlfix{red}}} \hlfix{column with a five-pointed star on top denotes the lowest value of 0.76.}}
	\label{fig:loss-hyper}
	\vspace{0pt}
\end{figure*}

\subsection{Ablation Study}
\label{sec:ablation-study}
\hlB{To better verify the robustness and effectiveness of RRL-Net, extensive ablation experiments on the VIS-NIR patch dataset are performed to explore the performance of each component.}

\hlB{\textbf{\textit{1) Break-down Ablation.}} We first explore the impact of each component on performance. From \cref{tab:Ablation-table-break-down}, the MGLA module, FIL module and MLPP optimization strategy can reduce the FPR95 by 33.3\%, 25.5\% and 24.8\%, respectively. These results verify the significant effectiveness of each component.}

\hlB{\textbf{\textit{2) Attention Scheme Comparison.}} For the cross-spectral image patch matching task, it is difficult to fully extract effective features by simply focusing on local features. Therefore, we propose a lightweight MGLA module to extract global features in multiple dimensions and strengthen the local feature interactions within global features. We compare our MGLA module with various attention modules on the VIS-NIR patch dataset. From \cref{tab:Ablation-table-MGLA}, firstly, adding an attention module can improve the matching results, and our MGLA module with negligible parameters achieves SOTA results. According to statistics, the number of parameters and computational complexity of the 10 MGLA modules in the encoder are only 120 and 0.004 GFLOPs. Secondly, from the results of the SE modules with different compression rates, excessive compression of the channel dimension has a negative impact, and the impact becomes more serious with the increase of compression rate. At the same time, combined results from the ECA, ELA, and MGLA modules indicate that avoiding dimensionality reduction while enhancing local interactions within global features significantly improves cross-spectral image patch matching performance. Finally, results from the SE and CBAM modules show that adding a spatial attention branch to emphasize local features does not improve matching performance and can even degrade it. At the same time, combining the results of the SNL, GC and MGLA modules, using attention pooling operations to more fully capture global features can significantly improve matching performance. In summary, for the cross-spectral image patch matching, on the one hand, we should pay more attention to the global features of individual image patches to fully exploit the discriminative features. On the other hand, we should strengthen the local interactions within the global features to effectively obtain the local dependencies.}

\hlB{To further explore the performance of the MGLA module, we split the MGLA module into a spatial dimension \textit{H} global-to-local attention (HGLA) module, a spatial dimension \textit{W} global-to-local attention (WGLA) module and a channel dimension \textit{C} global-to-local attention (CGLA) module. From \cref{tab:Ablation-table-MGLA} and \cref{fig:roc}, firstly, compared with not using the MGLA module, the results are significantly improved whether using a single-dimensional global-to-local module (HGLA, WGLA, CGLA) or a multi-dimensional global-to-local module (MGLA). This verifies the effectiveness of the global-to-local attention idea. Secondly, the effect of adding the CGLA module is significantly better than that of HGLA and WGLA. This shows that compared with paying attention to other dimensions, fully considering the global context and strengthening the local interaction between the global features of the channel dimension can achieve a more significant improvement in matching performance. In addition, compared with the CGLA module, the MGLA module can achieve better matching results with a negligible increase in parameters, which can more comprehensively promote effective feature extraction of image patches.}

\begin{figure}[!t]
	\centering
	\includegraphics[width=\columnwidth]{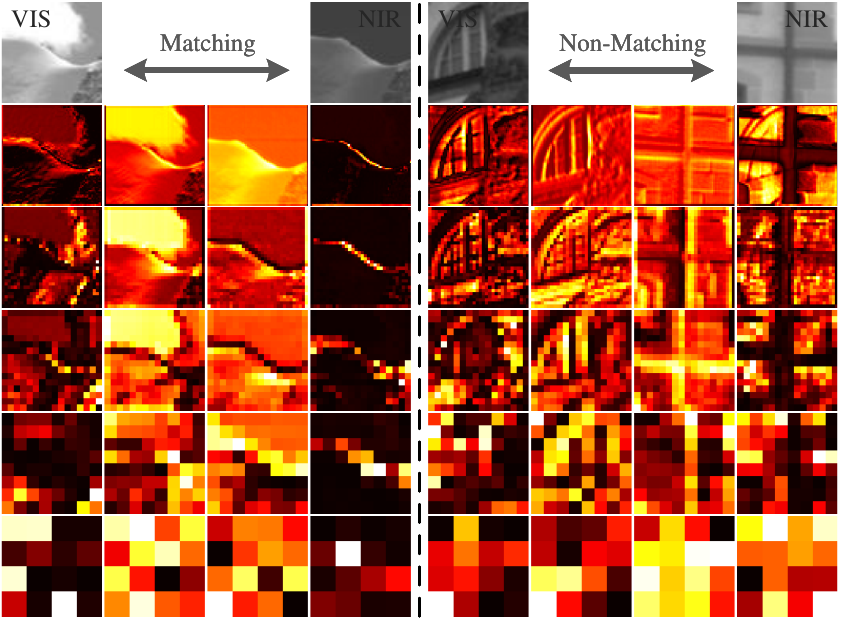}
	\vspace{-15pt}
	\caption{Visualization of the output feature maps of matching and non-matching VIS-NIR samples after passing through the MGLA modules and the FIL modules. For each pair of samples, the middle two columns denote the output of the MGLA module at each stage, and the left and right columns denote the private feature maps output by the FIL module at each stage. From top to bottom, they denote the stages from shallow to deep.}
	\label{fig:feature-visualization}
	\vspace{-8pt}
\end{figure}

\hlB{\textbf{\textit{3) ALFE network.}} Combined with the proposed MGLA module, we conduct an in-depth exploration of the feature extraction network. We explore the number of convolutional layers (before each MGLA module) and cross-layer connections at each stage.~\cite{yu2023feature} achieves the optimal matching result by using 3 convolutional layers and applying cross-layer connections. However, from \cref{tab:Ablation-table-ALFE}, when using our MGLA module, the optimal number of convolutional layers is 1 and no cross-layer connections are used. At the same time, compared with Variant 1-Variant 5, the FPR95 and parameters of the ALFE network are reduced by 5.0\% - 16.5\% and 1.0\% - 25.7\%. According to statistics, the number of parameters and computational complexity of the ALFE network are only 786,744 and 0.31 GFLOPs. These show that the addition of lightweight MGLA modules can reduce the use of network layers and cross-layer connections, and improve the network's ability to extract effective features of image patches.}

\hlB{\textbf{\textit{4) MLPP optimization strategy.}} To effectively verify the performance of our proposed MLPP strategy, we perform many experiments, as shown in \cref{tab:tab02}. Compared with not using the MLPP strategy, the reasonable addition of sub-metric networks and reconstruction branches will improve the matching results. At the same time, the improvement is more obvious when they are used superimposed, which can reduce the FPR95 by 24.8\%. Of course, it will bring about an increase in network parameters and inference time. To solve this problem, we use post-pruning. Compared with simply using the multi-loss strategy, the MLPP optimization strategy has the same matching effect improvement, and its network parameters and single-sample inference time are reduced by 26.7\% and 25.9\% (GTX 2080Ti), respectively. It is worth mentioning that compared with only adding sub-metric networks (``Variant 2''), which has achieved an excellent result of 0.83 in FPR95, adding additional reconstruction branches (``Variant 5'') will reduce FPR95 by 8.4\% (from 0.83 to 0.76). This proves that fully mining individual intrinsic features will help break the bottleneck of existing feature relation extraction, thereby achieving a significant improvement in matching performance. In addition, to verify the versatility of the MLPP strategy on different hardware platforms, we conducte the inference time test on different device (GTX 2080Ti and GTX 3090). From \cref{tab:tab02}, compared with Variant 1 - Variant 5, using the MLPP strategy can reduce single-sample inference time by 9.1\% - 25.9\% and 6.3\% - 30.2\% on GTX2080Ti and GTX 3090 devices. By comparing the reduction ratio on the two devices, the proposed MLPP strategy has a stable acceleration effect on different platforms.}

\hlB{At the same time, excessive constraints on the low-level features of the main network will cause it to focus on semantic feature extraction for sub-network discrimination and ignore the extraction of texture detail information for main network discrimination. Therefore, we conduct a detailed exploration of the position of the sub-metric branch to determine the reasonable combination of module losses and stage losses. From \cref{fig:loss-hyper}, when the added module loss and stage loss are both 3, RRL-Net has the lowest FPR95, which verifies that the reasonable addition of multiple sub-metric branches can effectively improve network performance.}

\hlB{In addition, we conduct experiments to explore the values of $\alpha$ and $\beta$, and the experimental results are shown in \cref{fig:loss-hyper}. First of all, when the value of $\alpha$ is within 0.1 - 10 and the value of $\beta$ is within 0.05 - 0.5, RRL-Net has good matching results. It shows that RRL-Net has a certain degree of stability. At the same time, when $\alpha$ is 1 and $\beta$ is 0.1, the network has the optimal matching result. In addition, when $\alpha$ is less than 1 or $\beta$ is less than 0.1 and they become smaller and smaller, FPR95 shows an upward trend. This is because the constraint of the reconstruction branch will be weakened, resulting in insufficient intrinsic feature mining. When $\alpha$ is greater than 1 or $\beta$ is greater than 0.1 and they become larger and larger, FPR95 also shows an upward trend and the trend is more obvious. The reason is that overemphasis on intrinsic feature mining and gradual neglect of subsequent feature relation mining will lead to insufficient discriminative feature extraction, resulting in a significant decrease in the matching effect. Notably, even in extreme cases (e.g., $\alpha = 100$ and $\beta=5$), the Mean FPR95 of the model on the VIS-NIR patch dataset is only 1.0. In addition, RRL-Net used the same parameter settings for experiments on all datasets and achieves significant and stable performance improvements, which can significantly reduce the burden of parameter adjustment on new datasets or tasks.}

\begin{table*}[t]
\caption{\hlfix{Stability and significance analysis of FIL-Net and RRL-Net on multiple datasets. \textbf{CI} denotes the confidence interval. \textbf{\textit{p}-value} is a significance indicator, which denotes the probability that the difference between the two groups occurs by chance.}}
    \vspace{-4pt}
	\label{tab:tab04}
	\centering
	\renewcommand{\arraystretch}{1.1}
	\setlength{\tabcolsep}{4mm}{
          \resizebox{\textwidth}{!}{

\begin{tabular}{c|ccc|ccc}
\hline
\multirow{2}{*}{\hlfix{Methods}} & \multicolumn{3}{c|}{\hlfix{VIS-NIR patch dataset}}                                                                               & \multicolumn{3}{c}{\hlfix{LLVIP patch dataset}}                                                                                  \\ \cline{2-7} 
                         & \multicolumn{1}{c|}{\hlfix{Mean ± Variance ↓}}      & \multicolumn{1}{c|}{\hlfix{95\% CI ↓}}                   & \hlfix{\textit{p}-value ↓}               & \multicolumn{1}{c|}{\hlfix{Mean ± Variance ↓}}      & \multicolumn{1}{c|}{\hlfix{95\% CI ↓}}                   & \hlfix{\textit{p}-value ↓}               \\ \hline
\hlfix{FIL-Net}                  & \multicolumn{1}{c|}{\hlfix{0.860 ± 0.050}}          & \multicolumn{1}{c|}{\hlfix{{[}0.799, 0.922{]}}}          & \multirow{2}{*}{\hlfix{0.0087}} & \multicolumn{1}{c|}{\hlfix{1.250 ± 0.062}}          & \multicolumn{1}{c|}{\hlfix{{[}1.173, 1.327{]}}}          & \multirow{2}{*}{\hlfix{0.0480}} \\ \cline{1-3} \cline{5-6}
\hlfix{RRL-Net}                  & \multicolumn{1}{c|}{\hlfix{\textbf{0.764 ± 0.038}}} & \multicolumn{1}{c|}{\hlfix{\textbf{{[}0.717, 0.811{]}}}} &                         & \multicolumn{1}{c|}{\hlfix{\textbf{1.160 ± 0.060}}} & \multicolumn{1}{c|}{\hlfix{\textbf{{[}1.086, 1.235{]}}}} &                         \\ \hline
\multirow{2}{*}{\hlfix{Methods}} & \multicolumn{3}{c|}{\hlfix{OS patch dataset}}                                                                                    & \multicolumn{3}{c}{\hlfix{SEN1-2 patch dataset}}                                                                                 \\ \cline{2-7} 
                         & \multicolumn{1}{c|}{\hlfix{Mean ± Variance ↓}}      & \multicolumn{1}{c|}{\hlfix{95\% CI ↓}}                   & \hlfix{\textit{p}-value ↓}               & \multicolumn{1}{c|}{\hlfix{Mean ± Variance ↓}}      & \multicolumn{1}{c|}{\hlfix{95\% CI ↓}}                   & \hlfix{\textit{p}-value ↓}               \\ \hline
\hlfix{FIL-Net}                  & \multicolumn{1}{c|}{\hlfix{5.366 ± 0.105}}          & \multicolumn{1}{c|}{\hlfix{{[}5.236, 5.496{]}}}          & \multirow{2}{*}{\hlfix{0.0425}} & \multicolumn{1}{c|}{\hlfix{1.540 ± 0.034}}          & \multicolumn{1}{c|}{\hlfix{{[}1.498, 1.582{]}}}          & \multirow{2}{*}{\hlfix{0.0003}} \\ \cline{1-3} \cline{5-6}
\hlfix{RRL-Net}                  &\multicolumn{1}{c|}{\hlfix{\textbf{5.222 ± 0.083}}}                      & \multicolumn{1}{c|}{\hlfix{\textbf{{[}5.120, 5.325{]}}}} &                         & \multicolumn{1}{c|}{\hlfix{\textbf{1.398 ± 0.041}}} & \multicolumn{1}{c|}{\hlfix{\textbf{{[}1.347, 1.449{]}}}} &                         \\ \hline
\end{tabular}
}}
\end{table*}


\begin{table}[t]
\caption{\hlfix{FPR95 comparison of various methods in the domain generalization experiment setting. \textbf{OS} denotes the OS patch dataset, \textbf{SEN1-2} denotes the SEN1-2 patch dataset, and \textbf{VIS-NIR} denotes the VIS-NIR patch dataset.}}
    \vspace{-4pt}\textbf{}
	\label{tab:tab05}
	\centering
	\renewcommand{\arraystretch}{1.2}
	\setlength{\tabcolsep}{3mm}{
          \resizebox{\columnwidth}{!}{
\begin{tabular}{ccc}
\hline
\hlfix{Methods} & \hlfix{OS → VIS-NIR}    & \hlfix{SEN1-2 → VIS-NIR} \\ \hline
\hlfix{AFD-Net\textsuperscript{\cite{quan2019afd,quan2021element}}} & \hlfix{17.17 (-16.1\%)} & \hlfix{7.15 (-15.7\%)}   \\
\hlfix{MR\_3A\textsuperscript{\cite{quan2021multi}}}  & \hlfix{20.27 (-29.0\%)} & \hlfix{8.09 (-18.0\%)}   \\
\hlfix{MFD-Net\textsuperscript{\cite{yu2022multibranch}}} & \hlfix{22.33 (-35.5\%)} & \hlfix{8.39 (-21.0\%)}   \\
\hlfix{EFR-Net\textsuperscript{\cite{yu2023efficient}}} & \hlfix{15.4 (-6.5\%)}   & \hlfix{7.27 (-8.8\%)}    \\
\hlfix{FIL-Net\textsuperscript{\cite{yu2023feature}}} & \hlfix{14.5 (-0.7\%))}  & \hlfix{8.88 (-25.3\%)}   \\ \hline
\rowcolor[HTML]{D7D7D7} 
\hlfix{RRL-Net (Ours)} & \hlfix{\textbf{14.4}}            & \hlfix{\textbf{6.63}}             \\ \hline
\end{tabular}
}}
\end{table}

\begin{figure}[t]
	\centering
	\includegraphics[width=\columnwidth]{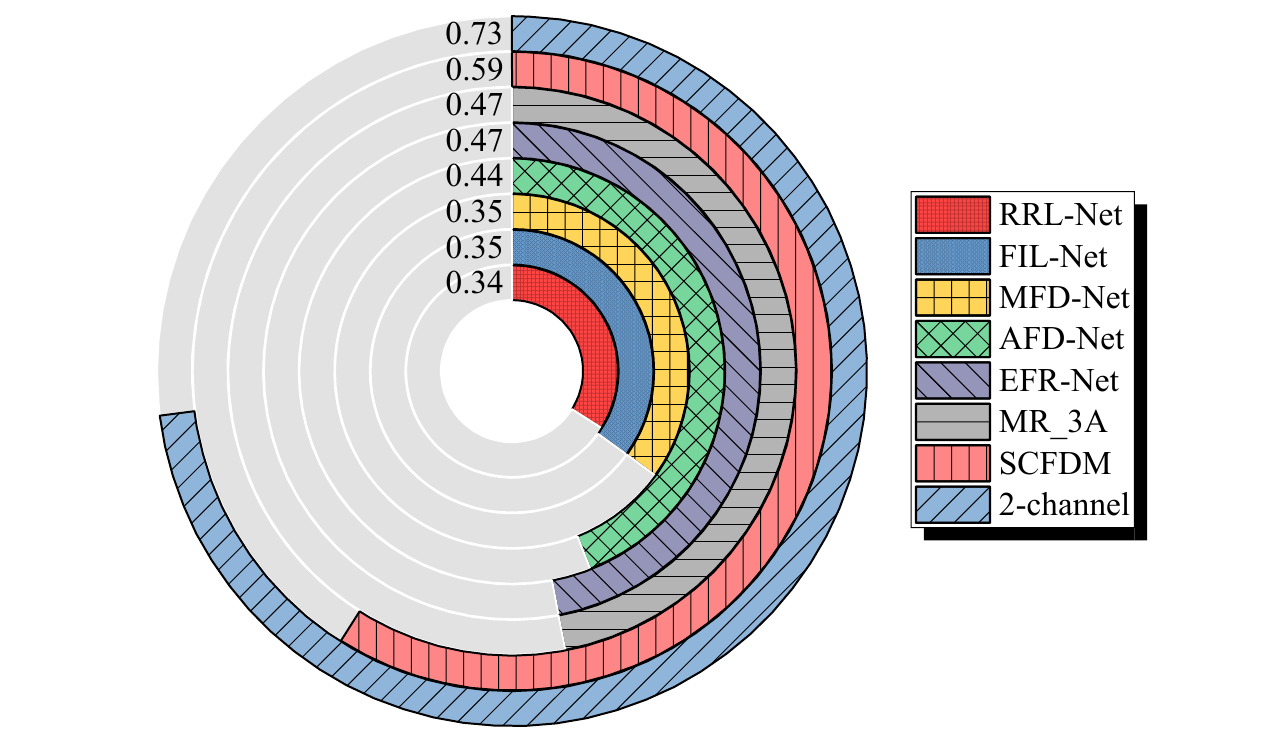}
	\vspace{-15pt}
	\caption{\hlfix{FPR95 comparison of various excellent image patch matching methods on the RIRE patch dataset. The RRL-Net achieves the best results.}}
	\label{fig:rire-methods-fpr95}
	\vspace{-8pt}
\end{figure}

\subsection{Visualization of Feature Mmaps} 
To further qualitatively analyze the performance of the MGLA module (ALFE Network) and FIL module in RRL-Net, we visualize the feature maps output by the corresponding modules for matching samples and non-matching samples as shown in \cref{fig:feature-visualization}. \hlfix{Bounded by the dotted line, a matching sample pair is shown on the left and a non-matching sample pair is shown on the right. The color brightness reflects the feature activation strength, which implies the degree of feature mining and regional response. By observing each sample pair in \cref{fig:feature-visualization}, the feature maps output by the MGLA module (the two middle columns) are significantly brighter than the feature maps output by the FIL module (the two side columns). This shows that the proposed MGLA model can fully characterize the intrinsic characteristics of the image patches. In addition, by observing the output of the FIL module of the two sample pairs, the brightness of the matching sample pairs is obviously lower, while that of the non-matching sample pairs is brighter. This phenomenon is due to the fact that the FIL module mines the differential features between image patches through feature interaction learning. For matching sample pairs, the private features between the two patches are relatively less. In contrast, non-matching sample pairs exhibit more significant differences, resulting in richer private features. This difference is further amplified by the subsequent metric branch, allowing RRL-Net to distinguish between matching and non-matching samples more accurately. This confirms the effectiveness of the FIL module in RRL-Net for capturing deep-level relations between image patches and enhancing matching discriminability.}

\begin{table*}[t]
\caption{\hlB{Performance comparison of different input image sizes on the VIS-NIR patch dataset.}}
\vspace{-4pt}
\label{tab:tab_image_size}
\centering
\renewcommand{\arraystretch}{1.1}
\setlength{\tabcolsep}{2.5mm}{
\resizebox{\textwidth}{!}{
\begin{tabular}{cccccccccc}
\hline
Input image size (pixel) & Field ↓         & Forest ↓        & Indoor ↓        & Mountain ↓      & Oldbuilding ↓   & Street ↓        & Urban ↓         & Water ↓         & Mean ↓          \\ \hline
48×48      & 4.43          & 0.05          & 1.52          & 1.77          & 0.84          & 0.56          & 0.32          & 2.44          & 1.49          \\
\rowcolor[HTML]{CFCECE} 
64×64      & 2.16          & 0.04          & 0.86          & 0.59          & 0.41          & 0.26          & 0.14          & 1.65          & 0.76          \\
80×80      & 1.50          & \textbf{0.02} & 0.74          & 0.48          & 0.45          & 0.25          & 0.20          & 1.30          & 0.62          \\
96×96      & \textbf{0.66} & 0.03          & \textbf{0.63} & \textbf{0.32} & \textbf{0.27} & \textbf{0.23} & \textbf{0.13} & \textbf{1.08} & \textbf{0.42} \\ \hline
\end{tabular}
}}
\end{table*}

\begin{figure*}[t]
	\centering
	\includegraphics[width=\textwidth]{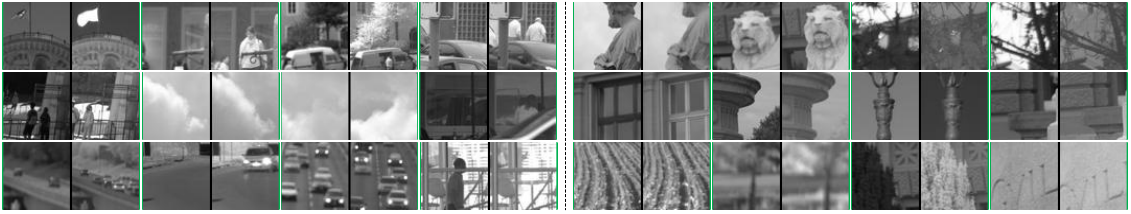}
	\vspace{-16pt}
	\caption{\hlB{Examples of some difficult matching pairs from the VIS-NIR patch dataset being correctly discriminated. The left side of the dotted line shows dynamic object shifts due to acquisition time differences, and the right side of the dotted line shows the local displacement phenomenon caused by the difference in sensor and imaging position.}}
	\label{fig:fig_ex_robustness_spatiotemporal}
	\vspace{-3pt}
\end{figure*}

\subsection{Analysis of Stability, Significance, and Generalization}

\hlfix{\textbf{\textit{1) Stability analysis.}} To further verify the stability of the proposed RRL-Net, we conduct 5 experiments on four datasets (VIS-NIR patch dataset, LLVIP patch dataset, OS patch dataset and SEN1-2 patch dataset) and statistically analyze the mean, variance and confidence intervals of the results. From \cref{tab:tab04}, RRL-Net achieves lower FPR95 mean on multiple datasets than that of FIL-Net. At the same time, except for the SEN1-2 patch dataset, RRL-Net exhibits significantly lower performance variance on the other three datasets. In addition, the 95\% CI of RRL-Net on the VIS-NIR patch dataset, OS patch dataset and SEN1-2 patch dataset are narrower. In summary, the proposed RRL-Net has excellent stability.}

\hlfix{\textbf{\textit{2) Significance analysis.}} To further verify the significance of the proposed RRL-Net, we perform significance analysis on FIL-Net and RRL-Net. From \cref{tab:tab04}, the \textit{p}-values on all datasets are less than 0.05, indicating that the performance improvement of RRL-Net compared to FIL-Net is statistically significant and not accidental. At the same time, compared with the LLVIP patch dataset and OS patch dataset, the \textit{p}-value on the VIS-NIR patch dataset and SEN1-2 patch dataset is smaller. The reason is that the VIS-NIR patch dataset and SEN1-2 patch dataset have more training samples, which makes the model training more sufficient, the results more stable, and RRL-Net can learn intrinsic features more effectively, thereby achieving more significant performance advantages.}

\hlfix{\textbf{\textit{3) Generalization analysis.}} To further verify the cross-scenario generalization ability of the proposed RRL-Net, we conduct a cross-domain generalization verification experiment: training on the VIS-SAR scene and testing on the VIS-NIR scene. Specifically, we use the OS patch dataset or SEN1-2 patch dataset belonging to the VIS-SAR scene for training, and test on the test set of the VIS-NIR patch dataset belonging to the VIS-NIR scene. From \cref{tab:tab05}, RRL-Net outperforms other methods in both cross-domain settings. Specifically, compared with other methods, RRL-Net reduces FPR95 by 0.7\% - 35.5\% in the “OS → VIS-NIR” setting, and reduces FPR95 by 8.8\% - 25.3\% in the “SEN1-2 → VIS-NIR” setting. At the same time, comparing the results of ``SEN → VIS-NIR'' and ``OS → VIS-NIR'' under the same method, the SEN1-2 patch dataset with more training samples as the training set have better generalization performance. This shows that sufficient data helps to enhance the generalization ability of the model. The above results jointly verify that RRL-Net exhibits superior robustness and good generalization ability when dealing with challenges such as spectral differences and sample domain changes.}

\subsection{Discussion}

\textbf{\textit{\hlB{1) Robustness of RRL-Net to image patch pairs with spatiotemporal differences.}}} \hlB{The proposed RRL-Net has anti-interference ability for sample pairs containing small displacement and noise. From \Cref{fig:fig_ex_robustness_spatiotemporal}, even when the VIS and NIR image patches are misaligned due to temporal differences (dynamic target position changes) or spatial differences (local offsets caused by sensor differences or imaging position deviations), RRL-Net can still correctly identify them as matching pairs. These results intuitively verify the robustness of the proposed network to spatiotemporal inconsistencies in real complex scenarios.}

\textbf{\textit{\hlB{2) The impact of input image size on matching performance.}}} \hlB{Our RRL-Net can directly adapt to input image patches of different sizes without modifying the network structure. To further verify the performance of RRL-Net under different patch sizes, we conduct additional experiments with 48×48 pixel, 80×80 pixel, and 96×96 pixel inputs, and compare the results with the default 64×64 pixel input. From \Cref{tab:tab_image_size}, when the input image patch size is larger, the model can achieve a lower FPR95 value. This is because larger image patches can provide more complete context and structural information, which helps to capture the global and local correspondences between cross-spectral image patches and reduce misjudgments. On the contrary, at a smaller patch size of 48×48 pixels, although the model still maintains strong discrimination ability, the performance is somewhat reduced compared to the 64×64 pixel patch due to insufficient available context information. Specifically, compared to the default 64×64 pixel image patch input, using 48×48 pixel patches increases the Mean FPR95 from 0.76 to 1.49, using 80×80 pixel patches reduces the Mean FPR95 from 0.76 to 0.62, and using 96×96 pixel patches reduces the Mean FPR95 from 0.76 to 0.42.}

\subsection{Application}
\hlfix{To further explore the performance of RRL-Net in various application scenarios, we apply it to the medical multi-modal image patch matching, cross-spectral image patch matching under high lighting differences, cross-spectral image registration and cross-spectral image fusion. The details are as follows:}

\hlfix{\textbf{\textit{1) Medical multi-modal image patch matching.}} To explore the application of RRL-Net in CT-MRI multi-modal scenarios, we compare it with various excellent image patch matching methods on the RIRE patch dataset. From \cref{fig:rire-methods-fpr95}, compared with other methods, the proposed RRL-Net drops by 2.9\% (from 0.35 to 0.34) - 53.4\% (from 0.73 to 0.34) on FPR95. At the same time, the FPR95 of RRL-Net on the RIRE patch dataset is only 0.34. The above results comprehensively verify the matching ability and cross-modal generalization performance of RRL-Net in medical tasks.}

\begin{figure}[t]
	\centering
	\includegraphics[width=\columnwidth]{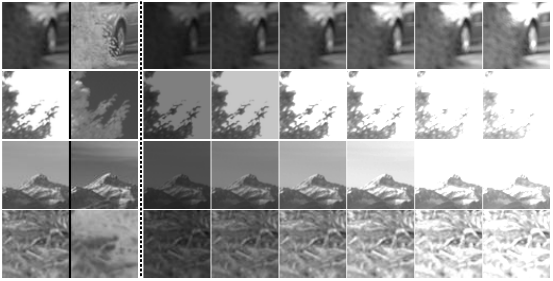}
	\vspace{-18pt}
	\caption{\hlfix{Visualization of different brightness adjustments of VIS image patches. The left and right sides of the dotted line are the VIS-NIR image pair and the VIS image patch under different brightness adjustments, respectively. The brightness adjustment range is [0.5, 2.0], and is presented every 0.3.}}
	\label{fig:brightness-adjustments}
	\vspace{-5pt}
\end{figure}

\textbf{\textit{\hlfix{2) Cross-spectral image patch matching under high lighting differences.}}} \hlfix{Although the LLVIP patch dataset covers a variety of lighting conditions, the acquisition time of its VIS and LWIR images is almost the same, so it only reflects the lighting changes between image pairs, and cannot reflect the matching problem of high lighting differences between two image patches within an image pair. To further study the impact of inconsistent lighting between image patches (such as day-night pairing) on matching performance, we design a simulation experiment based on the VIS-NIR patch dataset. Specifically, by artificially adjusting the brightness of VIS image patches, we construct scenes with different lighting intensities. From \cref{fig:brightness-adjustments}, as the brightness changes, the VIS image gradually becomes over-exposed or under-exposed, resulting in obvious degradation of texture and structural information. The nonlinear difference between it and the NIR image patch is significantly aggravated, and the matching task becomes more difficult. This experiment can provide an intuitive basis for analyzing the stability of cross-spectral matching under extreme lighting conditions. To further quantitatively evaluate the adaptability of the proposed RRL-Net in scenes with high lighting differences, we systematically compare it with FIL-Net under multiple sets of brightness adjustment settings. From \cref{tab:tab06}, RRL-Net significantly outperforms FIL-Net under all brightness adjustment settings with a lower FPR95, verifying its matching robustness under extreme lighting conditions. In addition, further analysis of the performance changes of RRL-Net under different brightness adjustment ranges reveals that although extreme lighting condition may lead to a certain degree of decline in matching performance, RRL-Net maintains stable performance within reasonable adjustments ([0.5, 2.0]), achieving an FPR95 of just 1.02 on the VIS-NIR patch dataset. The results show that the proposed RRL-Net exhibits good stability and robustness in cross-spectral matching tasks with significant lighting differences between image patches.}

\begin{table}[t]
\caption{\hlfix{Performance comparison between RRL-Net and FIL-Net under different brightness adjustment settings.}}
    \vspace{-4pt}
	\label{tab:tab06}
	\centering
	\renewcommand{\arraystretch}{1.1}
	\setlength{\tabcolsep}{4mm}{
          \resizebox{\columnwidth}{!}{
\begin{tabular}{ccc}
\hline
\hlfix{Brightness adjustment range} & \hlfix{FIL-Net} & \hlfix{RRL-Net} \\ \hline
\hlfix{{[}0.5 - 2.0{]}}             & \hlfix{1.09}    & \hlfix{1.02}    \\
\hlfix{{[}0.5 -1.5{]}}              & \hlfix{0.97}    & \hlfix{0.90}    \\
\hlfix{{[}0.7 -1.5{]}}              & \hlfix{0.92}    & \hlfix{0.83}    \\
\hlfix{{[}0.9 -1.2{]}}              & \hlfix{0.86}    & \hlfix{0.79}    \\ \hline
\rowcolor[HTML]{D7D7D7} 
\hlfix{No adjustment}               & \hlfix{\textbf{0.85}}    & \hlfix{\textbf{0.76}}    \\ \hline
\end{tabular}
}}
\end{table}

\begin{figure*}[tb]
\begin{minipage}{\textwidth}
	\centering
	\includegraphics[width=\textwidth]{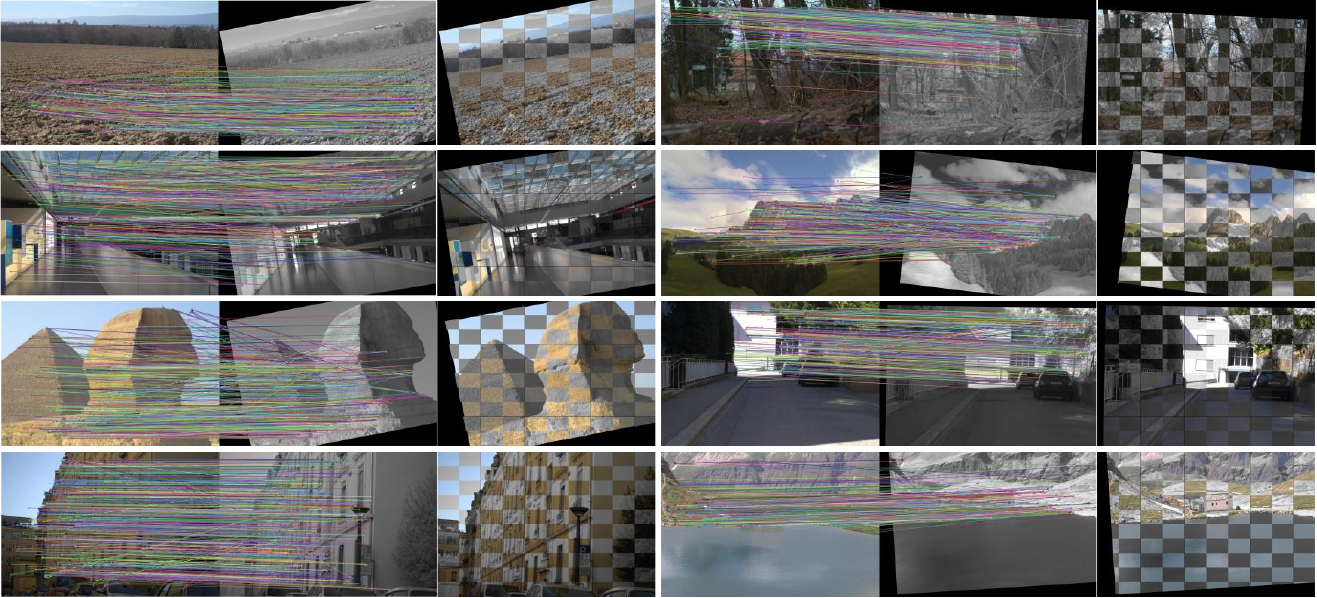}
	\vspace{-18pt}
	\caption{Partial registration results of RRL-Net on the VIS-NIR registration dataset.}
	\label{fig:registration-sample-display}
\end{minipage}

\vspace{10pt}

\begin{minipage}{\textwidth}
	\centering
	\includegraphics[width=\textwidth]{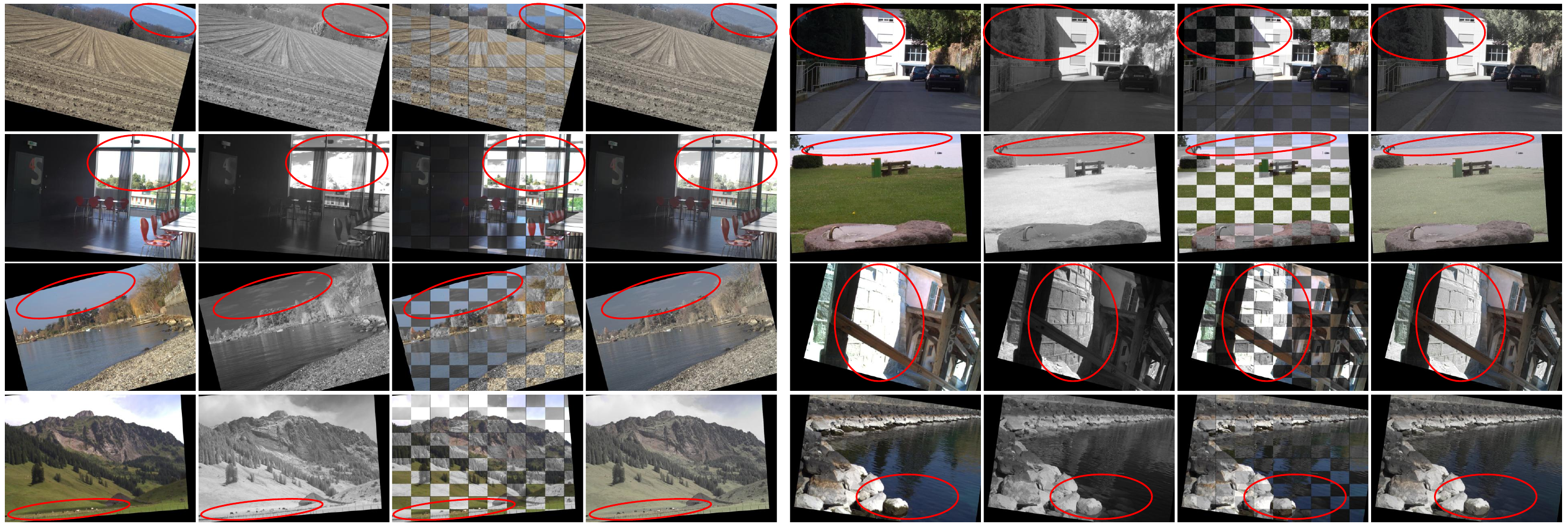}
	\vspace{-15pt}
	\caption{\hlfix{Image fusion effect on the VIS-NIR registration dataset. For each sample, the first column is the registered VIS image, the second column is the reference NIR image, the third column is the registered chessboard image, and the fourth column is the fused image.}}
	\label{fig:fusion-sample-display}
\end{minipage}
\vspace{-8pt}
\end{figure*}

\textbf{\textit{3) Cross-spectral image registration.}} To verify the robustness and generalization ability of RRL-Net, we apply our RRL-Net to the cross-spectral image registration task. Consistent with the setting of ~\cite{yu2023feature}, we first use the SIFT algorithm to extract 1000 keypoints in each image. Secondly, the image patches cropped around the keypoints are combined and input into the RRL-Net to measure the similarity. Then, a nearest neighbor search and threshold filtering are used to find the correspondence between the cross-spectral image patches. Finally, RANSAC~\cite{fischler1981random} is used to eliminate erroneous matches, thereby calculating the registration parameters and achieving registration. From \cref{tab:tab07}, although our RRL-Net achieves suboptimal results in registration rate, compared with the optimal registration rate, RRL-Net is only 0.6\% different from it. It is worth mentioning that the parameters of RRL-Net only account for 6\% of MFD-Net. At the same time, our RRL-Net achieves the best results on mRMSE and is less than 5. The above shows that our RRL-Net has excellent robustness in cross-spectral image registration. In addition, RRL-Net has the most registered samples with RMSE less than 1. This shows that RRL-Net has a more refined registration effect. We also display some registration results on the VIS-NIR registration dataset. From the checkerboard images in \cref{fig:registration-sample-display}, the VIS images after affine transformation almost completely overlap with the NIR images, which proves that RRL-Net has excellent registration results when applied to cross-spectral image registration.

\hlfix{\textbf{\textit{4) Cross-spectral image fusion.}} Different spectral images have complementary perceptual characteristics. Fusing multi-source information helps reduce the uncertainty in common areas and achieve a more comprehensive and accurate scene expression. Considering that the image pairs to be fused in practice are often not registered, we adopt a two-stage strategy of first registration and then fusion. In the image registration stage, we adopt a cross-spectral image registration method based on RRL-Net, which has robust feature modeling and relationship expression capabilities, and can effectively alleviate the impact of modal differences on matching accuracy. In the image fusion stage, we introduce the dual-scale image fusion (TIF) strategy~\cite{bavirisetti2016two} to simultaneously preserve structural contours and detailed textures, achieving higher quality cross-spectral fusion. Specifically, we conduct experiments on the VIS-NIR registration dataset. Consistent with the previous settings, the ``Country'' subset is used as the training set and the remaining subsets are used as the test set. From the red circled area in \cref{fig:fusion-sample-display}, due to the complex lighting conditions, a single VIS image is prone to overexposure (such as areas with direct strong light) and underexposure (such as areas blocked by shadows), which leads to serious loss of target detail information in local areas. In contrast, near-infrared images have a strong ability to resist light interference and can more stably capture the structure and contour information of objects. However, due to the absence of color information, it typically exhibits limitations in perception and scene discrimination. By observing the fused image, it can effectively make up for the structural defects of the VIS image in the unreasonable lighting area, while also make up for the lack of color in the NIR image. The overall results show that cross-spectral image fusion has significant advantages in improving the recognition of target areas and the consistency of structural expression. It not only improves the image quality, but also lays a more robust foundation for downstream tasks.}

\begin{table}[t]
	\scriptsize
	\caption{Registration results of various excellent image patch matching methods on the VIS-NIR registration dataset. \textbf{Bold} denotes optimal, \underline{underlined} denotes suboptimal.}
    \vspace{-4pt}
	\label{tab:tab07}
	\centering
	\renewcommand{\arraystretch}{1}
	\setlength{\tabcolsep}{1.5mm}{
          \resizebox{\columnwidth}{!}{
		\begin{tabular}{c|ccc|c|c}
			\toprule
			& \multicolumn{3}{c|}{\quad RMSE distribution \quad}       &                        &                           \\
			\multirow{-2}{*}{\raisebox{0.6ex}{\quad\quad Methods\quad\quad}} & \quad \textless 1 ↑  & \, 1$\sim$5 \,& \textgreater 5 ↓ \quad& \multirow{-2}{*}{\raisebox{0.6ex}{\quad $I\textsubscript{rr}$ ↑ \quad}} & \multirow{-2}{*}{\raisebox{0.6ex}{\, \textit{mRMSE} ↓ \,}} \\
			\midrule
			SCFDM\textsuperscript{\cite{quan2018cross}}            & 121           & 29       & 10               & 93.8\%                 & 8.48                      \\
			AFD-Net\textsuperscript{\cite{quan2019afd,quan2021element}}      & 121           & 29       & 10               & 93.8\%                 & 9.30                      \\
			MR\_3A\textsuperscript{\cite{quan2021multi}}            & 118           & 31       & 11               & 93.1\%                 & 10.43                     \\
			MFD-Net\textsuperscript{\cite{yu2022multibranch}}           & \underline{125}     & 30       & \textbf{5}       & \textbf{96.9\%}        & 5.57                      \\
			EFR-Net\textsuperscript{\cite{yu2023efficient}}           & 124           & 30       & \underline{6}          & \underline{96.3\%}           & \underline{4.95}                \\
			FIL-Net\textsuperscript{\cite{yu2023feature}}          & 115           & 38       & 7                & 95.6\%                 & 5.27                      \\
			\midrule
			\rowcolor[HTML]{E0E0E0} 
			RRL-Net (Ours)            & \textbf{127}  & 27       & \underline{6}          & \underline{96.3\%}           & \textbf{4.62}  \\
			\bottomrule          
		\end{tabular}
	}
    }
    \vspace{-10pt}
\end{table}

\section{Conclusion}
\label{Conclusion}
In this work, we innovatively construct \hlfix{a lightweight RRL-Net that addresses} the bottleneck of subsequent feature relation extraction caused by insufficient intrinsic feature \hlfix{representation} of individual image patches in existing methods. To solve the insufficient extraction of individual image patch intrinsic features, on the one hand, we propose a MGLA module and construct an ALFE network \hlfix{to enhance both global and local intrinsic feature representations.} On the other hand, we construct a feature reconstruction branch, which together with the ALFE network forms an autoencoder {to further strengthen intrinsic modeling. For the feature relations between image patches,} we use a FIL module to fully mine the rich and deep-level feature relations. In addition, we construct a MLPP optimization strategy to facilitate network optimization. \hlfix{The comparison results verify that RRL-Net achieves SOTA performance with superior robustness and wide applicability on multiple multi-modality datasets. Notably, this study focuses on algorithmic research on publicly available datasets and does not involve the collection or use of privacy information.}

\bibliographystyle{elsarticle-num} 
\bibliography{main}

\end{document}